\documentclass{article}

\usepackage[preprint]{neurips_2026}

\usepackage[utf8]{inputenc}
\usepackage[T1]{fontenc}
\usepackage[breaklinks=true,bookmarks=false]{hyperref}
\usepackage{url}
\usepackage{booktabs}
\usepackage{amsfonts}
\usepackage{amsmath}
\usepackage{amssymb}
\usepackage{nicefrac}
\usepackage{microtype}
\usepackage{xcolor}
\usepackage{graphicx}
\usepackage{placeins}
\usepackage{multirow}

\newcommand{\Erase}{\mathcal{E}}

\title{What Do EEG Foundation Models Capture from Human Brain Signals?}

\author{%
  \textbf{Ling Tang}\textsuperscript{1,2}\thanks{Equal contribution. Work done during an internship at Shanghai Artificial Intelligence Laboratory.}\quad
  \textbf{Qian Chen}\textsuperscript{1,4}\footnotemark[\value{footnote}]\quad
  \textbf{Jilin Mei}\textsuperscript{1,3}\quad
  \textbf{Houshi Xu}\textsuperscript{2}\quad
  \textbf{Quanshi Zhang}\textsuperscript{2}\\
  \textbf{Jing Shao}\textsuperscript{1}\quad
  \textbf{Na Zou}\textsuperscript{1,5}\quad
  \textbf{Xia Hu}\textsuperscript{1}\quad
  \textbf{Dongrui Liu}\textsuperscript{1}\thanks{Correspondence to: Dongrui Liu <liudongrui@pjlab.org.cn>}\\[4pt]
  \textsuperscript{1}Shanghai Artificial Intelligence Laboratory \quad
  \textsuperscript{2}Shanghai Jiao Tong University \\
  \textsuperscript{3}Fudan University \quad
  \textsuperscript{4}Tongji University \quad
  \textsuperscript{5}University of Houston
}

\begin{document}

\maketitle

\begin{abstract}
Clinical electroencephalogram (EEG) analysis rests on a hand-crafted feature catalog refined over decades,  \emph{e.g.,} band power, connectivity, complexity, and more. Modern EEG foundation models bypass this catalog, learn directly from raw signals via self-supervised pretraining, and match or outperform feature-engineered baselines on most clinical benchmarks. Whether the two representations align is an open question, which we decompose into three sub-questions: \emph{what does the model learn}, \emph{what does the model use}, and \emph{how much can be explained}. We answer them with layer-wise ridge probing, LEACE-style cross-covariance subspace erasure, and a transparent classifier benchmarked against a random-feature baseline. The audit covers three foundation models (CSBrain, CBraMod, LaBraM), five clinical tasks (MDD, Stress, ISRUC-Sleep, TUSL, Siena), and a 6-family 63-feature lexicon. Of the $945$ (model, task, feature) units, $648$ ($68.6\%$) are representation-causal and $199$ ($21.1\%$) are encoded-only. Across tasks, $50$ features qualify as universal candidates with strong support (all three architectures RC) in two or more tasks. Frequency-domain features dominate, but the other five families each contribute substantial causal mass. Confirmed features recover, on average, $79.3\%$ of the foundation model's advantage over the random baseline, with a clean task gradient (MDD $\approx 0.99$ down to Stress $\approx 0.56$): tasks near ceiling are almost fully recovered by the lexicon, while harder tasks leave a non-trivial residual that pinpoints a concrete target for future concept discovery. \\
\end{abstract}

\section{Introduction}
\label{sec:intro}

Electroencephalography (EEG) is a non-invasive technique that records the brain's electrical activity at the scalp, originating from the synchronous postsynaptic activity of cortical pyramidal neurons via volume conduction~\citep{kirschstein2009source, buzsaki2004neuronal}. For decades, clinicians have read these signals to diagnose disease: a neurologist marks epileptiform discharges to confirm epilepsy~\citep{smith2005eeg}, a sleep technician scores 30-second epochs, and a psychiatrist examines spectral patterns. These readings rest on a structured set of hand-crafted EEG features, grouped into six families: time-domain morphology, spectral power and shape, time-frequency envelope dynamics, signal complexity, cross-frequency coupling, and cross-channel relations.

In recent years, EEG foundation models~\citep{kostas2021bendr, jiang2024labram, zhou2025csbrain, wang2025cbramod, yue2024eegpt} have begun to change this picture. Pre-trained on large unlabeled EEG corpora via self-supervised learning, they achieve state-of-the-art performance on downstream tasks; on sleep staging, deep neural networks reach inter-rater agreement on par with expert scorers~\citep{stephansen2018sleep}. These results suggest the models have learned signal structures relevant to clinical decisions, possibly including ones the existing features do not name.

\begin{figure}[t]
\centering
\includegraphics[width=\linewidth]{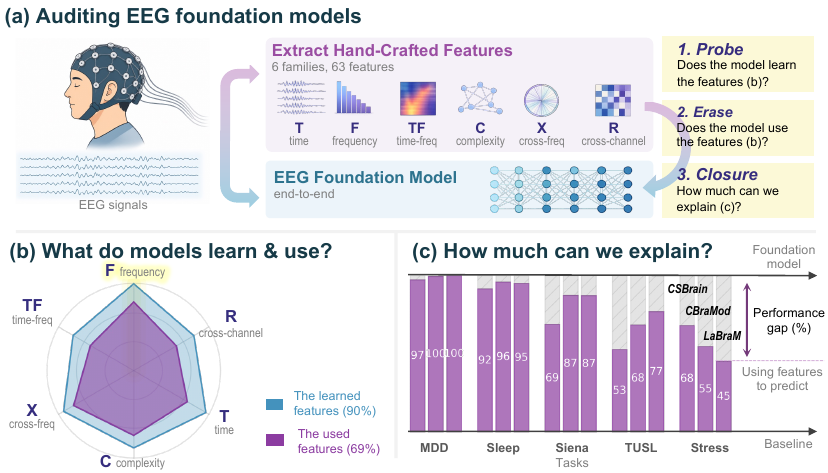}
\caption{Overview of the three-question audit. \textbf{(a)} A 6-family $63$-feature hand-crafted EEG lexicon is the audit target; for every (dataset, model, feature) triplet across three foundation models and five clinical tasks ($945$ units in total) we run \emph{Probe}, \emph{Erase} (LEACE-style cross-covariance subspace erasure), and \emph{Closure}. \textbf{(b)} Family-aggregated learn-vs-use rates across all $945$ cells: outer ring is encoded ($\approx 90\%$), inner ring is causally used ($\approx 69\%$). \textbf{(c)} Per-(task, model) closure ratios against a same-dimension random-feature baseline; closure follows a clean MDD-to-Stress task gradient.}
\label{fig:fig1}
\end{figure}

However, what the model has learned is hidden inside a high-dimensional representation. The core question is whether this representation aligns with the existing features or has diverged from them; either answer is informative. If aligned, identifying which features drive the model's decisions can validate or extend the existing set. If diverged, the model is using signal structures the existing features have not named, and identifying these structures would expand the catalogue of clinical EEG descriptors. Either outcome creates a feedback loop between model and human knowledge, a path we see as central to using foundation models in AI for Science.

We decompose this question into three sub-questions. \textit{\textbf{What does the model learn?}} Which hand-crafted features does the model encode, and at which layer? \textit{\textbf{What does the model use?}} Of the encoded features, which ones does the model rely on for task decisions? \textit{\textbf{How much can be explained?}} If we take the features the model uses and train a transparent classifier on them, how much of the foundation model's advantage over a same-dimension random-feature baseline can we recover? Together, the three measure alignment in encoding, in usage, and in performance recovery.

We answer each sub-question with a distinct interpretability method. Layer-wise ridge probing predicts each feature from the model's representation at every layer, revealing what is encoded but not what is used~\citep{belinkov2022probing, vig2020investigating}. A LEACE-style cross-covariance subspace erasure~\citep{belrose2023leace}, a scalable approximation of the closed-form LEACE operator, then removes the linearly decodable component of the target feature at its encoding peak; the resulting drop in held-out test-split task performance reveals causal reliance, with three null-target controls, a linear residual probe, and per-panel Benjamini--Hochberg correction ruling out spurious effects~\citep{ravfogel2020inlp}. Finally, a logistic-regression classifier trained on the confirmed features in each (dataset, model) cell yields a closure ratio against a random-feature baseline matched in dimension to the confirmed feature set. We assemble a lexicon of 6 families covering 63 features, and run all three methods on every (dataset, model, feature) triplet across three foundation models and five clinical tasks ($945$ units total). Figure~\ref{fig:fig1} previews the pipeline.

Across the $945$ units, the audit yields four findings. \textit{\textbf{(i) Encoding.}} All $63$ hand-crafted features are encoded in at least one (model, task) cell; $847$ of $945$ units ($89.6\%$) pass the encoding criterion, \emph{i.e.}, foundation models recover the canon broadly without prior. \textit{\textbf{(ii) Usage.}} Under held-out test FDR, $648$ of $945$ units ($68.6\%$) are representation-causal and $199$ ($21.1\%$) are encoded-only; across tasks, $50$ features are universal candidates with strong support in two or more tasks ($7$ task-specific, $6$ model-specific). \textit{\textbf{(iii) Family dominance.}} Frequency-domain features lead, but the other five families each contribute substantial causal mass on at least four of the five tasks. \textit{\textbf{(iv) Closure ceiling.}} Confirmed features recover $79.3\%$ of the foundation model's advantage over the random-feature baseline on average, with a clean task gradient (MDD $\approx 0.99$ down to Stress $\approx 0.56$): tasks near ceiling are almost fully recovered by the lexicon, while harder tasks leave a non-trivial residual that locates a concrete target for future concept-discovery research.

\section{Related work}
\label{sec:related}

Our work draws on three lines of research that have so far developed in parallel: hand-crafted EEG features in clinical and neuroscience research, EEG foundation models, and interpretability methods for EEG classifiers.

\textbf{Hand-crafted EEG features in clinical and neuroscience research.} Decades of clinical and neuroscience work have produced a rich catalog of EEG features. Time-domain morphology is captured by the Hjorth parameters~\citep{hjorth1970eeg}; frequency-domain content by band power across the canonical $\delta, \theta, \alpha, \beta, \gamma$ rhythms~\citep{welch1967fft, buzsaki2004neuronal}. Cross-frequency interactions have been characterized as mechanisms for neural communication~\citep{canolty2006high, canolty2010functional, jensen2007crossfrequency, lisman2013thetagamma}, with the Tort modulation index~\citep{tort2010measuring} the standard measure. Inter-channel relations are formalized through the phase-locking value~\citep{lachaux1999measuring} and the phase-lag index~\citep{stam2007phaselag}, signal complexity through permutation entropy~\citep{bandt2002permutation} and sample entropy~\citep{richman2000sample}, and surrogate-data tests support the family~\citep{theiler1992testing, schreiber2000surrogate, prichard1994generating}. The features examined in this paper are drawn entirely from this body of work; we contribute no new feature definitions, only a causal audit of which of these established features foundation models actually use.

\textbf{EEG foundation models.} Self-supervised pretraining on unlabeled EEG corpora has produced a family of foundation models~\citep{kostas2021bendr, wang2023brainbert, wang2023biot, wu2023brant, jiang2024labram, yue2024eegpt, foumani2024eeg2rep, cui2024neurogpt, jiang2024neurolm, zhou2025csbrain, wang2025cbramod}, spanning contrastive pretraining, masked patch reconstruction, long-context transformers, and cross-scale tokenization. These works report strong downstream performance, but which signal content their representations rely on has remained largely open.

\textbf{Interpretability for EEG models.} Existing work mainly applies post-hoc attribution to individual classifiers: SHAP and LRP for clinical classifiers~\citep{sylvester2024sherpa, nam2023lrp, yi2025lrpmdd}, Grad-CAM and SHAP for sleep staging and schizophrenia diagnosis~\citep{dutt2023sleepxai, saadatinia2024sczxai, srinivasan2025sczxai}, surrogate-based channel importance and ictal attribution~\citep{lee2025neuroxai, lapera2024ictalxai}, frequency-band perturbation~\citep{nahmias2020easypeasi}, and concept-activation vectors for transformers~\citep{skaaning2023tcav}. These methods report correlations between input features and predictions, but correlation alone does not establish causal use~\citep{belinkov2022probing}, and they are typically scoped to a single task and feature family.

We connect these three lines by bringing causal concept erasure from NLP~\citep{ravfogel2020inlp, ravfogel2022rlace, belrose2023leace} to EEG foundation models. Applied across three models, five clinical tasks, and a 6-family 63-feature lexicon, the audit separates what foundation models encode, what they use, and how much of their advantage that usage explains.

\section{Method}
\label{sec:method}

We address the three questions raised in Section~\ref{sec:intro} with three interpretability methods, applied to every (dataset, model, feature) triplet. The setup and notation are common to all three.

\subsection{Setup and notation}
\label{sec:method-setup}

We work with three pretrained EEG foundation models, $m \in \mathcal{M} = \{\text{CSBrain}, \text{CBraMod}, \text{LaBraM}\}$, on five clinical EEG tasks, $d \in \mathcal{D} = \{\text{MDD}, \text{Stress}, \text{Sleep}, \text{TUSL}, \text{Siena}\}$. Each foundation model has $L_m$ analyzable backbone layers (encoder layers for CSBrain and CBraMod, transformer blocks for LaBraM); the activation of model $m$ at layer $l$ on an EEG epoch $x \in \mathbb{R}^{C \times T}$ is $h_{m, l}(x) \in \mathbb{R}^{d_{m, l}}$. The pretrained model with linear classifier head is $f_m$, and $M(f_m, d)$ is its task metric on $d$ (ROC-AUC for binary tasks MDD, Stress, Siena; macro-F1 for multi-class tasks Sleep and TUSL).

The feature lexicon $\mathcal{Q}$ contains $63$ hand-crafted EEG features ($49$ per-channel and $14$ global) drawn from the literature surveyed in Section~\ref{sec:related} and grouped into six families covering time-domain morphology, spectral power and shape, time-frequency envelope dynamics, signal complexity, cross-frequency coupling, and cross-channel relations; the full registry with formal definitions is in Appendix~\ref{app:lexicon}. We write $z_q(x) \in \mathbb{R}^{p_q}$ for the value of feature $q \in \mathcal{Q}$ on $x$. The audit visits every $(d, m, q)$ triplet, $|\mathcal{D}| \times |\mathcal{M}| \times |\mathcal{Q}| = 945$ in total.

All datasets are split into training, validation, and test partitions. The training partition supplies all standardization, probe fitting, eraser estimation, residual probe fitting, and transparent-classifier training; the validation partition supplies regularization strength, the encoding peak layer, and the selection-side encoded flag; the test partition supplies all reported metrics and the held-out confirmation used by the representation-causal criterion below.

\subsection{What does the model learn?}
\label{sec:method-learn}

To determine whether the foundation model's representation contains the information of each hand-crafted feature, we ask whether a linear function of an activation can predict the feature's value, following the probing-classifier paradigm of language and vision interpretability~\citep{conneau2018probing, belinkov2022probing}. We use ridge regression as the probe family (most hand-crafted EEG features are continuous-valued), and we fit one probe per backbone layer to localize where each feature lives along the depth of the network.

For each triplet $(d, m, q)$ and each layer $l \in \{1, \ldots, L_m\}$, we fit a probe
\begin{equation}
P_{d, m, q, l} \,:\; h_{m, l}(x) \;\mapsto\; z_q(x),
\label{eq:probe}
\end{equation}
on the training partition with regularization chosen on validation. The probe strength on split $s \in \{\mathrm{val}, \mathrm{test}\}$ is the coefficient of determination clipped at zero,
\begin{equation}
R^{(s)}_{d, m, q, l} \;=\; \max\!\bigl(0,\; R^{2}_{s}\bigl(P_{d, m, q, l}(h_{m, l}(x)),\; z_q(x)\bigr)\bigr),
\label{eq:probe-strength}
\end{equation}
averaged across output dimensions when $p_q > 1$. The encoding peak layer is selected on the validation split, $l^{*}_{d, m, q} = \arg\max_{l} R^{(\mathrm{val})}_{d, m, q, l}$, and the test-split probe strength $R^{(\mathrm{test})}_{d, m, q, l^{*}}$ is reported as held-out evidence. We say feature $q$ is \emph{selection-encoded} on $(d, m)$ if its validation-split probe strength exceeds a pre-registered threshold and exceeds matched shuffled-target and Gaussian-target controls by a pre-registered margin (Appendix~\ref{app:protocol-encoding}); the validation flag determines whether the triplet enters the erasure stage. A separate test-side encoded flag, based on the analogous test-split criterion, is reported for held-out auditing.

\subsection{What does the model use?}
\label{sec:method-use}

Linear readability is a property of the activation, not of the network's downstream computation~\citep{belinkov2022probing, vig2020investigating}, so to separate use from encoding we intervene on the activation. We use a LEACE-style cross-covariance subspace erasure~\citep{belrose2023leace}, a scalable variant of the line of causal concept erasure methods~\citep{ravfogel2020inlp, ravfogel2022rlace, ravfogel2022kernel}: we center the activation and the target on the training partition, form their empirical cross-covariance, and remove the retained left-singular-vector subspace from the centered activation. This is a covariance-simplified approximation rather than the full covariance-whitened minimum-distortion LEACE operator (see Appendix~\ref{app:reproducibility} for the explicit relation), and we revisit this approximation in the limitations of Section~\ref{sec:discussion}.

At the encoding peak $l^{*} \equiv l^{*}_{d, m, q}$, write $h \equiv h_{m, l^{*}}(x)$, $\mu_h = \mathbb{E}[h]$, $\mu_z = \mathbb{E}[z_q]$, and $\Sigma_{hz} = \mathbb{E}[(h-\mu_h)(z_q-\mu_z)^{\top}]$. We center activations and targets on the training partition; we do not estimate or use $\Sigma_{hh}$, so the eraser is the Euclidean orthogonal projector onto the column span of $\Sigma_{hz}$ rather than the whitened LEACE projector. We form $\Sigma_{hz} = U S V^{\top}$ by singular value decomposition and let $U_r$ collect the left singular vectors whose singular values exceed a fixed ridge threshold (or, if none do, the leading $p_q$ vectors). The eraser is
\begin{equation}
\Erase_{d, m, q}(h) \;=\; h - (h - \mu_h) \,\Pi_q,
\qquad
\Pi_q \;=\; U_r U_r^{\top},
\label{eq:leace}
\end{equation}
the projection onto the target-correlated subspace within the activation. We construct an edited model $\tilde{f}_{m, q}$ that runs the standard forward pass up to $l^{*}$, replaces $h_{m, l^{*}}(x)$ with $\Erase_{d, m, q}(h_{m, l^{*}}(x))$, and continues with the unchanged remainder of the network and classifier head. No model parameters are modified. The drop in test-split task metric on the edited model gives the erasure contribution,
\begin{equation}
\Delta_{\mathrm{erase}}(d, m, q) \;=\; M(f_m, d) - M(\tilde{f}_{m, q}, d).
\label{eq:delta-erase}
\end{equation}

A drop alone is not sufficient: it could arise from any same-dimension perturbation, a misaligned input--target pairing, or an arbitrary same-dimensional target. We therefore add three null-target controls~\citep{ravfogel2020inlp, ravfogel2022rlace} and a linear residual probe: (i) a random orthogonal subspace whose dimension matches the target expansion dimension $p_q$ (not necessarily the retained eraser rank), (ii) a shuffled pairing $(x_i, z_q(x_i))$, (iii) i.i.d.\ Gaussian noise of matching dimension, and (iv) a ridge probe refit on $\Erase_{d, m, q}(h)$. We call $q$ \emph{representation-causal} on $(d, m)$ when four conditions hold: the validation-side selection-encoded flag is true, the test-split bootstrap lower bound for $\Delta_{\mathrm{erase}}$ exceeds $0$, the per-panel BH/FDR $q$-value on the test split is below $0.05$, and $\Delta_{\mathrm{erase}} - \Delta_{\mathrm{erase}}^{\mathrm{(rand)}} > 0$. The remaining controls and the residual probe are reported as audit fields.

\subsection{How much can be explained?}
\label{sec:method-explain}

The third question closes the loop: if the confirmed features can stand in for the model on its task, the audit is performance-faithful, in the spirit of concept bottleneck models~\citep{koh2020concept}. For each cell $(d, m)$, let $\mathcal{Q}^{*}_{d, m} = \{\,q \in \mathcal{Q} : q \text{ is representation-causal on } (d, m)\,\}$, and train a logistic-regression classifier $c_{d, m}$ on the concatenation $\bigl[z_q(x)\bigr]_{q \in \mathcal{Q}^{*}_{d, m}}$ with a fixed $L_{2}$ penalty (Appendix~\ref{app:reproducibility}). As a floor, $B_{\mathrm{rand}}$ is the same classifier on i.i.d.\ Gaussian features matched in dimension to $\mathcal{Q}^{*}_{d, m}$, isolating feature \emph{identity} from feature \emph{count}. The closure ratio is
\begin{equation}
\mathrm{Closure}(d, m) \;=\; \frac{M(c_{d, m}, d) - M(B_{\mathrm{rand}}, d)}{M(f_m, d) - M(B_{\mathrm{rand}}, d) + \epsilon},
\label{eq:closure}
\end{equation}
the fraction of $f_m$'s advantage over $B_{\mathrm{rand}}$ that the confirmed features recover; here $B_{\mathrm{rand}}$ is fitted per cell so that its column count matches $|\mathcal{Q}^{*}_{d, m}|$, and $\epsilon = 10^{-12}$ guards the denominator on cells where $|M(f_m, d) - M(B_{\mathrm{rand}}, d)|$ is numerically zero (Appendix~\ref{app:protocol-closure}). A minimal-spectrum baseline $B_0$ (per-channel canonical-band log FFT-bin energies, F001--F005, expanded to $30$ columns under the aggregation rule of Appendix~\ref{app:protocol-feature-matrix}) is reported as a reference but does not enter Equation~\ref{eq:closure}.

\section{Experiments}
\label{sec:exp}

\begin{figure}[!t]
\centering
\includegraphics[width=\linewidth]{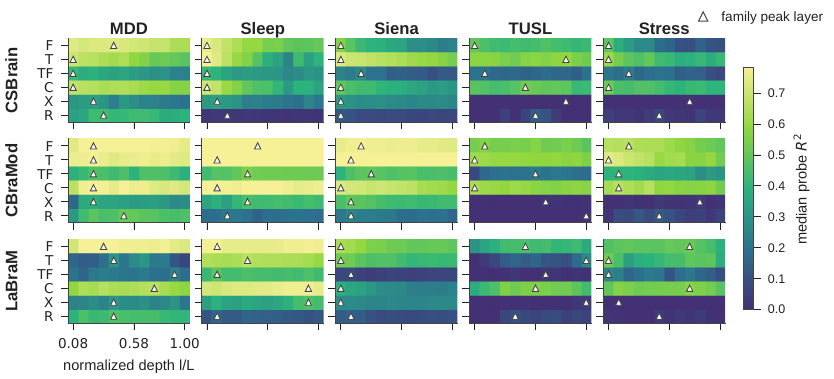}
\caption{Encoding atlas. Median probe $R^{2}$ over the features in each family, as a function of normalized depth $l / L_m$, for each of the $15$ $(d, m)$ cells (rows: $5$ tasks; columns: CSBrain, CBraMod, LaBraM). Triangles mark the family encoding peak layer. Frequency-domain features are encoded with high probe strength on every cell; time-frequency, cross-channel, and complexity families are encoded with moderate strength on most cells.}
\label{fig:probe-atlas}
\end{figure}

\subsection{Setup}
\label{sec:exp-setup}

\textbf{Datasets and models.} The audit runs on five EEG clinical tasks, \emph{i.e.}, MDD on Mumtaz, Stress on Mental Arithmetic, ISRUC-Sleep, TUSL seizure-type, and Siena seizure-detection, scored by ROC-AUC for binary tasks and macro-F1 for multi-class. The three foundation models are CSBrain~\citep{zhou2025csbrain}, CBraMod~\citep{wang2025cbramod}, and LaBraM~\citep{jiang2024labram}; for each $(d, m)$ cell we load the released fine-tuned checkpoint with a linear classifier head and freeze the backbone. The lexicon is the 6-family $63$-feature catalog of Appendix~\ref{app:lexicon}, expanded to a $328$-dimensional audit row.

\textbf{Splits and statistics.} We use each dataset's published subject-level train/validation/test partition, so no subject leaks across partitions; train fits every probe, eraser, residual probe, and transparent classifier, validation selects the probe regularization and encoding peak layer, and test reports all metrics. Paired bootstraps use $128$ resamples on the analysis row (sample for MDD/Stress/TUSL/Siena; $30$\,s epoch for Sleep) with shared resample indices, and per-panel Benjamini--Hochberg FDR controls multiple comparisons across the $63$ features at $q<0.05$. The full protocol is in Appendix~\ref{app:protocol}.

\subsection{Foundation models encode most of the lexicon}
\label{sec:exp-learn}

This experiment tests whether each hand-crafted feature is encoded in the foundation model's representation, and at what depth. For each $(d, m, q)$ triplet we fit a ridge regression probe at every backbone layer of $f_m$, score it on validation and test by the coefficient of determination clipped at zero (Section~\ref{sec:method-learn}), and record the encoding peak layer from the validation score. Probes share the train/val/test partition with the foundation model's classifier head; multi-dimensional features are scored by averaging $R^{2}$ over output dimensions.

The feature matrix is well-formed on every dataset (no non-finite entries on any of the $328$ expansion columns); only one feature, the Higuchi fractal dimension $C004$, triggers a low-variance quality-control flag on four of the five datasets and is kept with a starred annotation. Of the $945$ triplets, $847$ ($89.6\%$) reach the validation encoding criterion, and all $63$ features pass the criterion in at least one $(d, m)$ cell. Encoding strength varies smoothly with feature family and with model depth (Figure~\ref{fig:probe-atlas}). The frequency-domain family $F$ is encoded across every cell with median probe $R^{2}$ between $0.7$ and $0.85$; the time-domain ($T$) and complexity ($C$) families are also broadly encoded; cross-frequency coupling ($X$) and cross-channel relations ($R$) are encoded at moderate strength on most cells. Encoding peak layers cluster in the lower-to-middle backbone depths for CSBrain and CBraMod and span a wider depth range for LaBraM blocks; the same feature can sit at different depths in the three architectures, which means the LEACE intervention site in the next experiment is feature- and model-specific rather than fixed.

Foundation models pretrained on raw EEG, then, do recover most of the literature's hand-crafted features in their internal representation. This is consistent with the broader observation that self-supervised pretraining produces representations rich in human-interpretable structure, and it sets up the next question: whether the model actually uses what it has encoded.
\FloatBarrier

\subsection{Causal usage is broadly shared with task-local emphases}
\label{sec:exp-use}

Encoding does not imply usage. To test whether $f_m$ relies on each encoded feature for its task decisions, we apply the LEACE-style cross-covariance subspace eraser at the feature's encoding peak layer, propagate the edited activation through the unchanged remainder of the network, and measure the held-out test-split drop $\Delta_{\mathrm{erase}}$ in task metric (Section~\ref{sec:method-use}). Three null-target controls (random subspace, shuffled target, Gaussian same-dim) and a linear residual probe rule out drops driven by anything other than the target feature itself, and per-panel Benjamini--Hochberg correction guards against multiple-comparison false positives.

Of the $945$ triplets, $648$ ($68.6\%$) pass the test-supported representation-causal criterion, $199$ ($21.1\%$) remain encoded-only, and $98$ ($10.4\%$) are not-encoded. The pattern of representation-causal features is broadly shared across tasks, with task-local emphases in which family carries the largest individual erasures (full per-cell breakdown in Appendix~\ref{app:robustness}, Table~\ref{tab:erasure-family}). On TUSL/LaBraM, several time-frequency features (TF006 wavelet variance L5 and the TF007--TF008 envelope CV in $\delta/\theta$) and the related cross-frequency features (X005, X002) approach a metric-saturation drop of $\Delta_{\mathrm{erase}} \approx 0.529$. On MDD/CBraMod, cross-channel relations lead by family-level median ($\Delta_{\mathrm{erase}} = 0.081$ over $13$ representation-causal features), with frequency-domain features close behind by both count ($16$) and median ($0.076$). On Sleep, frequency-band relative powers and ratios carry the largest family-level effects on CSBrain (median $0.043$) and CBraMod (median $0.304$), while on LaBraM the complexity family $C$ edges out $F$ (median $0.210$ vs $0.183$). Of the $15$ features with the largest cross-cell maximum erasure (Appendix~\ref{app:robustness}, Table~\ref{tab:erasure-top20}), most are causally relied upon in all five tasks, and $6$ tie at the maximum drop $0.529$. The same probe and erasure pipeline succeeds on every cell, including Stress, where every model retains a non-empty $\mathcal{Q}^{*}_{d, m}$.
\FloatBarrier

\begin{figure}[!t]
\centering
\includegraphics[width=\linewidth]{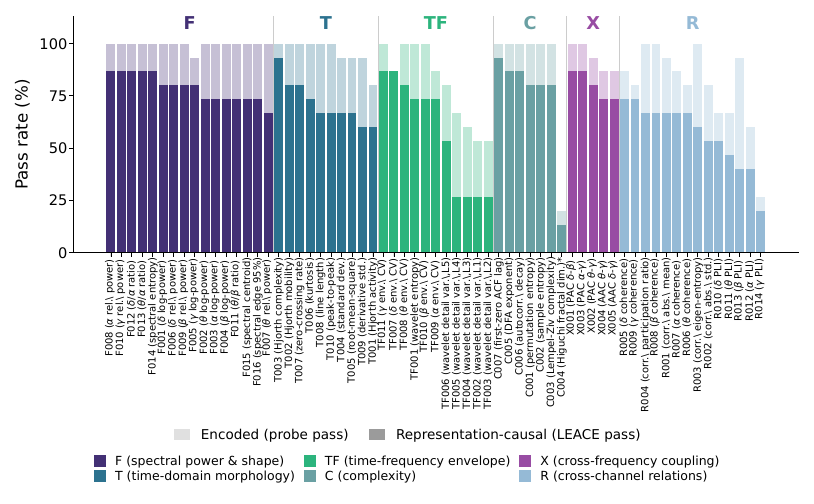}
\caption{Per-feature encoded vs.\ causal-use rate across the $15$ $(d, m)$ cells. Each bar is one of the $63$ hand-crafted features, grouped by feature family (top labels: F, T, TF, C, X, R; family colors). The light-shaded outer bar is the encoded rate (probe pass), the saturated inner bar is the causally-used rate (LEACE pass under held-out test FDR). Frequency-domain features (Family $F$) are encoded almost everywhere and used most consistently; the only feature with an encoded rate $\ll 1$ across all cells is the Higuchi fractal dimension $C004$, kept with a starred annotation.}
\label{fig:taxonomy}
\end{figure}
\FloatBarrier

Figure~\ref{fig:taxonomy} reports the encoded vs.\ causal-use rate for each individual feature across the $15$ $(d, m)$ cells; the gap between the light (encoded) and dark (causally used) bar is the encoded-only population. Aggregating these per-feature rates into cross-task categories, where a feature has \emph{strong task-level support} when all three architectures agree as representation-causal on the same task, $50$ features qualify as \emph{universal candidates} (strong support in two or more tasks), $7$ are \emph{task-specific} (strong support in exactly one task), and $6$ are \emph{model-specific} (no task reaches strong support, but at least one $(d, m)$ cell is representation-causal). No feature is encoded-only at the cross-task level. Feature usage is therefore broad: $50/63$ of the lexicon is relied upon by all three architectures on at least two tasks.

\begin{figure}[!t]
\centering
\includegraphics[width=\linewidth]{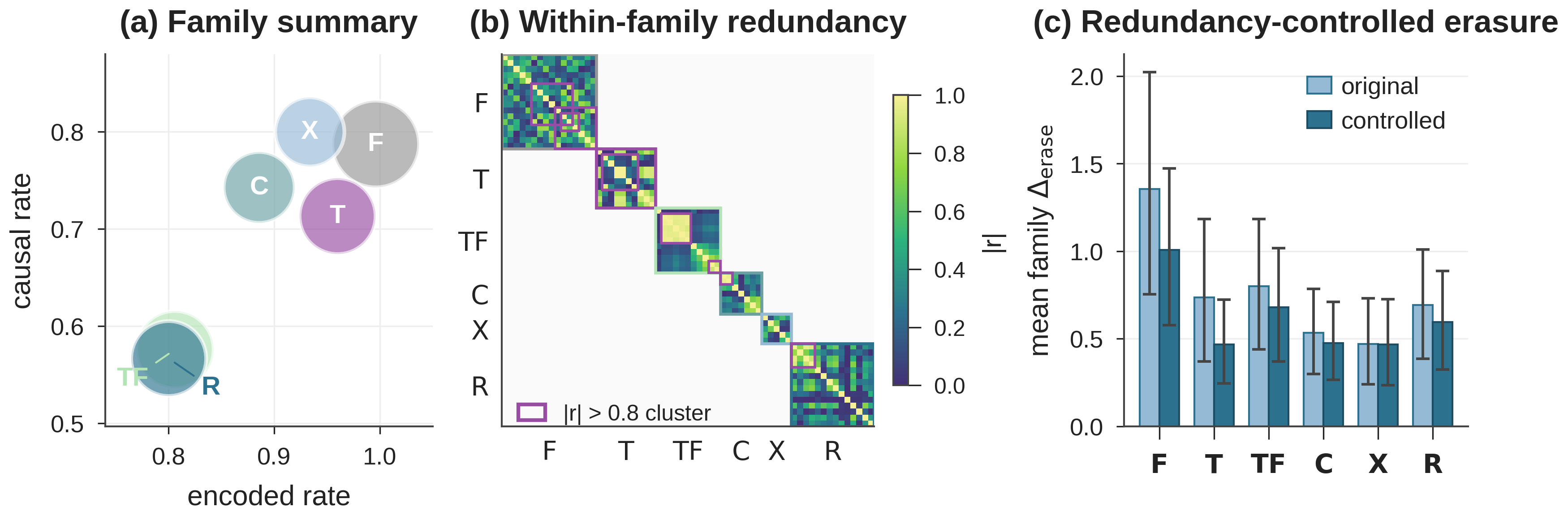}
\caption{Family-level summary. (A) Encoded rate vs.\ causal rate per family, with circle area proportional to cumulative effect mass. (B) Within-family $|r|$ matrix; red boxes mark $|r| > 0.8$ redundant clusters. (C) Mean family-level $\Delta_{\mathrm{erase}}$ before and after collapsing redundant clusters; the controlled effect tracks the original.}
\label{fig:family}
\end{figure}

At the family level, frequency-domain features remain dominant on encoded rate and cumulative effect mass, but the time-frequency, cross-channel, complexity, time-domain, and cross-frequency families each contribute substantial causal mass on at least four of the five tasks (Figure~\ref{fig:family}A). Within $F$, several $|r| > 0.8$ clusters of band-power and ratio features appear (Figure~\ref{fig:family}B); when we re-erase each family with redundant features collapsed to a single representative, the family-level $\Delta_{\mathrm{erase}}$ does not change appreciably (Figure~\ref{fig:family}C), so the causal signal in $F$ is not driven by counting near-duplicate features. The combination of broad cross-task usage, family-level dominance of frequency-domain features, and substantial contributions from the remaining five families together form, to our reading, the most actionable observation for the EEG XAI community: any probing-only interpretability claim about EEG foundation models should be paired with an intervention before being read as evidence of model reliance, and any conclusion that frequency-domain features alone explain the model should be tempered by the substantial mass carried by time-frequency, cross-channel, and complexity families on harder tasks.

\begin{table}[!t]
\centering
\caption{Closure on each $(d, m)$ pair, grouped by foundation model. Columns are the seven transparent classifiers and the foundation model: $B_0$ minimal-spectrum baseline, $B_{\mathrm{all}}$ all $63$ features, $B_{\mathrm{enc}}$ encoded features, $B_{\mathrm{rep}}$ representation-causal features (test-supported), $B_{\mathrm{fam}}$ family-supported features, $B_{\mathrm{rand}}$ same-dimension random-feature baseline (closure floor), and $\mathrm{FM}$ the foundation model itself. \emph{RC} is the number of representation-causal features and \emph{Closure} is $(B_{\mathrm{rep}}-B_{\mathrm{rand}})/(\mathrm{FM}-B_{\mathrm{rand}})$ from Equation~\ref{eq:closure}.}
\label{tab:closure}
\small
\setlength{\tabcolsep}{6pt}
\renewcommand{\arraystretch}{1.20}
\begin{tabular}{llccccccccc}
\toprule
Model & Task & $B_0$ & $B_{\mathrm{all}}$ & $B_{\mathrm{enc}}$ & $B_{\mathrm{rep}}$ & $B_{\mathrm{fam}}$ & $B_{\mathrm{rand}}$ & $\mathrm{FM}$ & RC & Closure \\
\midrule
\multirow{5}{*}{\textbf{CSBrain}}
       & MDD    & 0.981 & 0.985 & 0.986 & 0.984 & 0.985 & 0.477 & 0.998 & 37 & 0.974 \\
       & Sleep  & 0.661 & 0.752 & 0.750 & 0.741 & 0.752 & 0.191 & 0.792 & 47 & 0.915 \\
       & Siena  & 0.881 & 0.843 & 0.840 & 0.817 & 0.843 & 0.512 & 0.955 & 42 & 0.688 \\
       & TUSL   & 0.693 & 0.630 & 0.642 & 0.618 & 0.630 & 0.268 & 0.932 & 45 & 0.527 \\
       & Stress & 0.686 & 0.706 & 0.699 & 0.693 & 0.764 & 0.481 & 0.794 &  6 & 0.680 \\
\midrule
\multirow{5}{*}{\textbf{CBraMod}}
       & MDD    & 0.981 & 0.985 & 0.986 & 0.985 & 0.985 & 0.523 & 0.987 & 53 & 0.995 \\
       & Sleep  & 0.661 & 0.752 & 0.754 & 0.754 & 0.752 & 0.179 & 0.779 & 57 & 0.959 \\
       & Siena  & 0.881 & 0.843 & 0.840 & 0.864 & 0.843 & 0.488 & 0.918 & 32 & 0.874 \\
       & TUSL   & 0.693 & 0.630 & 0.637 & 0.597 & 0.630 & 0.231 & 0.766 & 47 & 0.685 \\
       & Stress & 0.686 & 0.706 & 0.721 & 0.723 & 0.706 & 0.530 & 0.884 & 47 & 0.547 \\
\midrule
\multirow{5}{*}{\textbf{LaBraM}}
       & MDD    & 0.981 & 0.985 & 0.986 & 0.989 & 0.985 & 0.516 & 0.984 & 47 & 1.012 \\
       & Sleep  & 0.661 & 0.752 & 0.750 & 0.750 & 0.752 & 0.190 & 0.778 & 54 & 0.951 \\
       & Siena  & 0.881 & 0.843 & 0.840 & 0.840 & 0.843 & 0.504 & 0.889 & 63 & 0.872 \\
       & TUSL   & 0.693 & 0.630 & 0.643 & 0.630 & 0.630 & 0.398 & 0.700 & 42 & 0.770 \\
       & Stress & 0.686 & 0.706 & 0.694 & 0.644 & 0.706 & 0.515 & 0.800 & 29 & 0.451 \\
\midrule
\multicolumn{10}{r}{Mean closure across the $15$ cells:} & \textbf{0.793} \\
\bottomrule
\end{tabular}
\end{table}
\FloatBarrier

\subsection{Confirmed features close most of the foundation-model gap, with a task gradient}
\label{sec:exp-explain}

The first two experiments identify what the model encodes and what it uses. The third asks whether the union of those used features can recover the foundation model's task advantage on its own. For each $(d, m)$ cell, we train a logistic-regression classifier on $\mathcal{Q}^{*}_{d, m}$, the set of representation-causal features confirmed by the previous step, and compare its task metric against a same-dimension random-feature baseline $B_{\mathrm{rand}}$ and against the foundation model. The closure ratio in Equation~\ref{eq:closure} reports the fraction of the foundation-model--baseline gap that the classifier recovers.

Across the $15$ $(d, m)$ cells, confirmed features recover $79.3\%$ of the foundation-model advantage over $B_{\mathrm{rand}}$ on average, with a clear task gradient (Table~\ref{tab:closure}). MDD is almost fully recovered on every architecture (closure $0.974$, $0.995$, $1.012$); Sleep follows ($0.915$, $0.959$, $0.951$); Siena sits in the middle ($0.688$, $0.874$, $0.872$); TUSL is partially recovered ($0.527$, $0.685$, $0.770$); Stress retains the largest residual ($0.680$, $0.547$, $0.451$). The largest closures coincide with the cells where the foundation model is closer to ceiling, and the smallest with the harder seizure-type and stress-detection tasks where individual erasures are also smaller in magnitude. The MDD/LaBraM closure of $1.012$ is a paired-bootstrap fluctuation in which the surrogate classifier matches the foundation model up to test-split noise; we report it as is rather than clipping at one. The Stress/CSBrain cell stands out for its small representation-causal set ($\mathrm{RC}=6$): with only six confirmed features it nonetheless recovers $0.680$ of the foundation-model advantage, consistent with the FDR-controlled criterion isolating a small but informative set on this architecture--task pair rather than indicating a pipeline failure.

The closure ratio is partial on the harder tasks: on TUSL and Stress, roughly $25\%$ to $55\%$ of the foundation-model advantage over $B_{\mathrm{rand}}$ lies outside our 6-family $63$-feature lexicon. Foundation models, while using human knowledge broadly, also rely on signal structures the existing features have not yet named or on non-linear combinations the linear surrogate cannot capture, which we discuss in Section~\ref{sec:discussion}.
\FloatBarrier

\subsection{Robustness and sensitivity}
\label{sec:exp-robustness}

\textbf{Failure record and redundancy control.} Of the $945$ triplets, $98$ fail the validation encoding criterion and are flagged not-encoded; only $C004$ (Higuchi fractal dimension) triggers a low-variance flag and is kept with a starred annotation. The closure pipeline passes a leakage audit end-to-end on all $15$ cells. Within-family redundancy is controlled by collapsing each $|r|>0.8$ cluster (most visible in Family $F$ band-power/ratio features) to its largest-$\Delta_{\mathrm{erase}}$ representative and re-aggregating; the redundancy-controlled family effect tracks the original within sampling noise on every cell, so family-level dominance of $F$ is not driven by near-duplicate counting.

\textbf{Sensitivity to family scope.} Three pre-registered conditions probe whether the closure ratio depends on which features enter $B_{\mathrm{rep}}$. \emph{Leave-one-family-out} changes the closure ratio by only a few percentage points on most cells, with TUSL and Stress most sensitive to removing the time-frequency or frequency families. \emph{Matched-dimension control} (a random feature subset of identical cardinality) lowers closure by $0.05$--$0.15$ on every cell, so the signal is feature-specific rather than dimension-driven. \emph{Top-$K$ per family} ($K=5$) tightens $B_{\mathrm{rep}}$ but leaves headline closure within a few percentage points. Together these conditions show that the task gradient in Table~\ref{tab:closure} is not an artefact of feature count or of any single family.

\section{Discussion and conclusion}
\label{sec:discussion}

EEG foundation models, trained without any inductive bias toward hand-crafted features, recover the clinical feature catalog at scale: $648$ of $945$ (model, task, feature) units ($68.6\%$) pass the representation-causal criterion, and confirmed features recover on average $79.3\%$ of the foundation-model advantage over a random-feature baseline, with a clean task gradient (MDD $\approx 0.99$ down to Stress $\approx 0.56$). The residual on harder tasks points to three targets for future concept discovery: descriptors outside the registry, non-linear feature combinations beyond a linear surrogate, and ceiling-bounded gaps on imbalanced or small-sample tasks.

\textbf{Limitations and broader impact.} Both probe and eraser are linear, restricting causal reliance to the linearly decodable component. The eraser is a cross-covariance approximation to LEACE (Appendix~\ref{app:reproducibility}); the four-condition criterion is evidence about this approximation, not the closed-form estimator. Several lexicon entries are GPU-friendly proxies for textbook estimators (Appendix~\ref{app:lexicon}). The lexicon is not exhaustive, and findings extend only to the three models and five tasks tested. The protocol generalizes to any domain pairing a feature catalog with a self-supervised foundation model; we analyze already-released models, and indirect over-trust risks are exactly what the encoded-vs-used distinction surfaces.

\bibliographystyle{plainnat}
\bibliography{references}

@article{hjorth1970eeg,
  author  = {Hjorth, Bo},
  title   = {{EEG} Analysis Based on Time Domain Properties},
  journal = {Electroencephalography and Clinical Neurophysiology},
  year    = {1970},
  volume  = {29},
  number  = {3},
  pages   = {306--310},
  doi     = {10.1016/0013-4694(70)90143-4}
}

@article{welch1967fft,
  author  = {Welch, Peter D.},
  title   = {The Use of Fast {F}ourier Transform for the Estimation of Power Spectra: A Method Based on Time Averaging Over Short, Modified Periodograms},
  journal = {{IEEE} Transactions on Audio and Electroacoustics},
  year    = {1967},
  volume  = {15},
  number  = {2},
  pages   = {70--73},
  doi     = {10.1109/TAU.1967.1161901}
}

@article{stam2007phaselag,
  author  = {Stam, Cornelis J. and Nolte, Guido and Daffertshofer, Andreas},
  title   = {Phase Lag Index: Assessment of Functional Connectivity from Multi Channel {EEG} and {MEG} with Diminished Bias from Common Sources},
  journal = {Human Brain Mapping},
  year    = {2007},
  volume  = {28},
  number  = {11},
  pages   = {1178--1193},
  doi     = {10.1002/hbm.20346}
}

@article{bandt2002permutation,
  author  = {Bandt, Christoph and Pompe, Bernd},
  title   = {Permutation Entropy: A Natural Complexity Measure for Time Series},
  journal = {Physical Review Letters},
  year    = {2002},
  volume  = {88},
  number  = {17},
  pages   = {174102},
  doi     = {10.1103/PhysRevLett.88.174102}
}

@article{richman2000sample,
  author  = {Richman, Joshua S. and Moorman, J. Randall},
  title   = {Physiological Time-series Analysis Using Approximate Entropy and Sample Entropy},
  journal = {American Journal of Physiology-Heart and Circulatory Physiology},
  year    = {2000},
  volume  = {278},
  number  = {6},
  pages   = {H2039--H2049},
  doi     = {10.1152/ajpheart.2000.278.6.H2039}
}

@article{kirschstein2009source,
  author  = {Kirschstein, Timo and K{\"o}hling, R{\"u}diger},
  title   = {What is the Source of the {EEG}?},
  journal = {Clinical {EEG} and Neuroscience},
  year    = {2009},
  volume  = {40},
  number  = {3},
  pages   = {146--149},
  doi     = {10.1177/155005940904000305}
}

@article{smith2005eeg,
  author  = {Smith, S. J. M.},
  title   = {{EEG} in the Diagnosis, Classification, and Management of Patients with Epilepsy},
  journal = {Journal of Neurology, Neurosurgery {\&} Psychiatry},
  year    = {2005},
  volume  = {76},
  number  = {Suppl 2},
  pages   = {ii2--ii7},
  doi     = {10.1136/jnnp.2005.069245}
}

@article{stephansen2018sleep,
  author  = {Stephansen, Jens B. and Olesen, Alexander N. and Olsen, Mads and Ambati, Aditya and Leary, Eileen B. and Moore, Hyatt E. and Carrillo, Oscar and Lin, Ling and Han, Fang and Yan, Han and Sun, Yun L. and Dauvilliers, Yves and Scholz, Sabine and Barateau, Lucie and H{\"o}gl, Birgit and Stefani, Ambra and Hong, Seung Chul and Kim, Tae Won and Pizza, Fabio and Plazzi, Giuseppe and Vandi, Stefano and Antelmi, Elena and Perrin, Dimitri and Kuna, Samuel T. and Schweitzer, Paula K. and Kushida, Clete and Peppard, Paul E. and S{\o}rensen, Helge B. D. and Jennum, Poul and Mignot, Emmanuel},
  title   = {Neural Network Analysis of Sleep Stages Enables Efficient Diagnosis of Narcolepsy},
  journal = {Nature Communications},
  year    = {2018},
  volume  = {9},
  number  = {1},
  pages   = {5229},
  doi     = {10.1038/s41467-018-07229-3}
}

@article{buzsaki2004neuronal,
  author  = {Buzs{\'a}ki, Gy{\"o}rgy and Draguhn, Andreas},
  title   = {Neuronal Oscillations in Cortical Networks},
  journal = {Science},
  year    = {2004},
  volume  = {304},
  number  = {5679},
  pages   = {1926--1929},
  doi     = {10.1126/science.1099745}
}

@article{canolty2006high,
  author  = {Canolty, Ryan T. and Edwards, Erik and Dalal, Sarang S. and Soltani, Maryam and Nagarajan, Srikantan S. and Kirsch, Heidi E. and Berger, Mitchel S. and Barbaro, Nicholas M. and Knight, Robert T.},
  title   = {High Gamma Power Is Phase-Locked to Theta Oscillations in Human Neocortex},
  journal = {Science},
  year    = {2006},
  volume  = {313},
  number  = {5793},
  pages   = {1626--1628},
  doi     = {10.1126/science.1128115}
}

@article{canolty2010functional,
  author  = {Canolty, Ryan T. and Knight, Robert T.},
  title   = {The Functional Role of Cross-Frequency Coupling},
  journal = {Trends in Cognitive Sciences},
  year    = {2010},
  volume  = {14},
  number  = {11},
  pages   = {506--515},
  doi     = {10.1016/j.tics.2010.09.001}
}

@article{jensen2007crossfrequency,
  author  = {Jensen, Ole and Colgin, Laura L.},
  title   = {Cross-Frequency Coupling between Neuronal Oscillations},
  journal = {Trends in Cognitive Sciences},
  year    = {2007},
  volume  = {11},
  number  = {7},
  pages   = {267--269},
  doi     = {10.1016/j.tics.2007.05.003}
}

@article{lachaux1999measuring,
  author  = {Lachaux, Jean-Philippe and Rodriguez, Eugenio and Martinerie, Jacques and Varela, Francisco J.},
  title   = {Measuring Phase Synchrony in Brain Signals},
  journal = {Human Brain Mapping},
  year    = {1999},
  volume  = {8},
  number  = {4},
  pages   = {194--208},
  doi     = {10.1002/(SICI)1097-0193(1999)8:4<194::AID-HBM4>3.0.CO;2-C}
}

@article{lisman2013thetagamma,
  author  = {Lisman, John E. and Jensen, Ole},
  title   = {The Theta-Gamma Neural Code},
  journal = {Neuron},
  year    = {2013},
  volume  = {77},
  number  = {6},
  pages   = {1002--1016},
  doi     = {10.1016/j.neuron.2013.03.007}
}

@article{prichard1994generating,
  author  = {Prichard, Dean and Theiler, James},
  title   = {Generating Surrogate Data for Time Series with Several Simultaneously Measured Variables},
  journal = {Physical Review Letters},
  year    = {1994},
  volume  = {73},
  number  = {7},
  pages   = {951--954},
  doi     = {10.1103/PhysRevLett.73.951}
}

@article{schreiber2000surrogate,
  author  = {Schreiber, Thomas and Schmitz, Andreas},
  title   = {Surrogate Time Series},
  journal = {Physica D: Nonlinear Phenomena},
  year    = {2000},
  volume  = {142},
  number  = {3--4},
  pages   = {346--382},
  doi     = {10.1016/S0167-2789(00)00043-9}
}

@article{theiler1992testing,
  author  = {Theiler, James and Eubank, Stephen and Longtin, Andr{\'e} and Galdrikian, Bryan and Farmer, J. Doyne},
  title   = {Testing for Nonlinearity in Time Series: The Method of Surrogate Data},
  journal = {Physica D: Nonlinear Phenomena},
  year    = {1992},
  volume  = {58},
  number  = {1--4},
  pages   = {77--94},
  doi     = {10.1016/0167-2789(92)90102-S}
}

@article{tort2010measuring,
  author  = {Tort, Adriano B. L. and Komorowski, Robert and Eichenbaum, Howard and Kopell, Nancy},
  title   = {Measuring Phase-Amplitude Coupling between Neuronal Oscillations of Different Frequencies},
  journal = {Journal of Neurophysiology},
  year    = {2010},
  volume  = {104},
  number  = {2},
  pages   = {1195--1210},
  doi     = {10.1152/jn.00106.2010}
}

@inproceedings{cui2024neurogpt,
  author        = {Cui, Wenhui and Jeong, Woojae and Th{\"o}lke, Philipp and Medani, Takfarinas and Jerbi, Karim and Joshi, Anand A. and Leahy, Richard M.},
  title         = {{Neuro-GPT}: Towards a Foundation Model for {EEG}},
  booktitle     = {2024 IEEE International Symposium on Biomedical Imaging (ISBI)},
  year          = {2024},
  archivePrefix = {arXiv},
  eprint        = {2311.03764},
  url           = {https://arxiv.org/abs/2311.03764}
}

@inproceedings{foumani2024eeg2rep,
  author        = {Foumani, Navid Mohammadi and Mackellar, Geoffrey and Ghane, Soheila and Irtza, Saad and Nguyen, Nam and Salehi, Mahsa},
  title         = {{EEG2Rep}: Enhancing Self-supervised {EEG} Representation through Informative Masked Inputs},
  booktitle     = {Proceedings of the 30th ACM SIGKDD Conference on Knowledge Discovery and Data Mining (KDD)},
  year          = {2024},
  archivePrefix = {arXiv},
  eprint        = {2402.17772},
  url           = {https://arxiv.org/abs/2402.17772}
}

@inproceedings{jiang2024labram,
  author        = {Jiang, Wei-Bang and Zhao, Li-Ming and Lu, Bao-Liang},
  title         = {{LaBraM}: Large Brain Model for Learning Generic Representations with Tremendous {EEG} Data in {BCI}},
  booktitle     = {The Twelfth International Conference on Learning Representations (ICLR)},
  year          = {2024},
  archivePrefix = {arXiv},
  eprint        = {2405.18765},
  url           = {https://openreview.net/forum?id=QzTpTRVtrP}
}

@inproceedings{jiang2024neurolm,
  author        = {Jiang, Wei-Bang and Wang, Yansen and Lu, Bao-Liang and Li, Dongsheng},
  title         = {{NeuroLM}: A Universal Multi-task Foundation Model for Bridging the Gap between Language and {EEG} Signals},
  booktitle     = {The Thirteenth International Conference on Learning Representations (ICLR)},
  year          = {2025},
  archivePrefix = {arXiv},
  eprint        = {2409.00101},
  url           = {https://openreview.net/forum?id=Io9yFt7XH7}
}

@article{kostas2021bendr,
  author        = {Kostas, Demetres and Aroca-Ouellette, Stephane and Rudzicz, Frank},
  title         = {{BENDR}: Using Transformers and a Contrastive Self-Supervised Learning Task to Learn from Massive Amounts of {EEG} Data},
  journal       = {Frontiers in Human Neuroscience},
  year          = {2021},
  volume        = {15},
  pages         = {653659},
  doi           = {10.3389/fnhum.2021.653659},
  archivePrefix = {arXiv},
  eprint        = {2101.12037},
  url           = {https://www.frontiersin.org/articles/10.3389/fnhum.2021.653659/full}
}

@inproceedings{wang2023biot,
  author        = {Yang, Chaoqi and Westover, M. Brandon and Sun, Jimeng},
  title         = {{BIOT}: Biosignal Transformer for Cross-data Learning in the Wild},
  booktitle     = {Advances in Neural Information Processing Systems 36 (NeurIPS)},
  year          = {2023},
  archivePrefix = {arXiv},
  eprint        = {2305.10351},
  url           = {https://openreview.net/forum?id=c2LZyTyddi}
}

@inproceedings{wang2023brainbert,
  author        = {Wang, Christopher and Subramaniam, Vighnesh and Yaari, Adam Uri and Kreiman, Gabriel and Katz, Boris and Cases, Ignacio and Barbu, Andrei},
  title         = {{BrainBERT}: Self-supervised Representation Learning for Intracranial Recordings},
  booktitle     = {The Eleventh International Conference on Learning Representations (ICLR)},
  year          = {2023},
  archivePrefix = {arXiv},
  eprint        = {2302.14367},
  url           = {https://openreview.net/forum?id=xmcYx_reUn6}
}

@inproceedings{wang2025cbramod,
  author        = {Wang, Jiquan and Zhao, Sha and Luo, Zhiling and Zhou, Yangxuan and Jiang, Haiteng and Li, Shijian and Li, Tao and Pan, Gang},
  title         = {{CBraMod}: A Criss-Cross Brain Foundation Model for {EEG} Decoding},
  booktitle     = {The Thirteenth International Conference on Learning Representations (ICLR)},
  year          = {2025},
  archivePrefix = {arXiv},
  eprint        = {2412.07236},
  url           = {https://openreview.net/forum?id=NPNUHgHF2w}
}

@inproceedings{wu2023brant,
  author    = {Zhang, Daoze and Yuan, Zhizhang and Yang, Yang and Chen, Junru and Wang, Jingjing and Li, Yafeng},
  title     = {Brant: Foundation Model for Intracranial Neural Signal},
  booktitle = {Advances in Neural Information Processing Systems 36 (NeurIPS)},
  year      = {2023},
  url       = {https://openreview.net/forum?id=DDkl9vaJyE}
}

@inproceedings{yue2024eegpt,
  author    = {Wang, Guangyu and Liu, Wenchao and He, Yuhong and Xu, Cong and Ma, Lin and Li, Haifeng},
  title     = {{EEGPT}: Pretrained Transformer for Universal and Reliable Representation of {EEG} Signals},
  booktitle = {Advances in Neural Information Processing Systems 37 (NeurIPS)},
  year      = {2024},
  url       = {https://openreview.net/forum?id=lvS2b8CjG5}
}

@inproceedings{zhou2025csbrain,
  author        = {Zhou, Yuchen and Wu, Jiamin and Ren, Zichen and Yao, Zhouheng and Lu, Weiheng and Peng, Kunyu and Zheng, Qihao and Song, Chunfeng and Ouyang, Wanli and Gou, Chao},
  title         = {{CSBrain}: A Cross-scale Spatiotemporal Brain Foundation Model for {EEG} Decoding},
  booktitle     = {Advances in Neural Information Processing Systems 38 (NeurIPS)},
  year          = {2025},
  archivePrefix = {arXiv},
  eprint        = {2506.23075},
  url           = {https://openreview.net/forum?id=agcXjEHmyW}
}

@article{belinkov2022probing,
  author    = {Belinkov, Yonatan},
  title     = {Probing Classifiers: Promises, Shortcomings, and Advances},
  journal   = {Computational Linguistics},
  year      = {2022},
  volume    = {48},
  number    = {1},
  pages     = {207--219},
  publisher = {MIT Press},
  doi       = {10.1162/coli_a_00422},
  url       = {https://aclanthology.org/2022.cl-1.7/}
}

@inproceedings{conneau2018probing,
  author    = {Conneau, Alexis and Kruszewski, German and Lample, Guillaume and Barrault, Lo{\"i}c and Baroni, Marco},
  title     = {What you can cram into a single $\$\&!\#*$ vector: {P}robing sentence embeddings for linguistic properties},
  booktitle = {Proceedings of the 56th Annual Meeting of the Association for Computational Linguistics (ACL)},
  year      = {2018},
  pages     = {2126--2136},
  doi       = {10.18653/v1/P18-1198},
  url       = {https://aclanthology.org/P18-1198/}
}

@inproceedings{belrose2023leace,
  author        = {Belrose, Nora and Schneider-Joseph, David and Ravfogel, Shauli and Cotterell, Ryan and Raff, Edward and Biderman, Stella},
  title         = {{LEACE}: Perfect Linear Concept Erasure in Closed Form},
  booktitle     = {Advances in Neural Information Processing Systems 36 (NeurIPS)},
  year          = {2023},
  archivePrefix = {arXiv},
  eprint        = {2306.03819},
  url           = {https://arxiv.org/abs/2306.03819}
}

@inproceedings{koh2020concept,
  author        = {Koh, Pang Wei and Nguyen, Thao and Tang, Yew Siang and Mussmann, Stephen and Pierson, Emma and Kim, Been and Liang, Percy},
  title         = {Concept Bottleneck Models},
  booktitle     = {Proceedings of the 37th International Conference on Machine Learning (ICML)},
  year          = {2020},
  pages         = {5338--5348},
  archivePrefix = {arXiv},
  eprint        = {2007.04612},
  url           = {https://proceedings.mlr.press/v119/koh20a.html}
}

@inproceedings{ravfogel2020inlp,
  author        = {Ravfogel, Shauli and Elazar, Yanai and Gonen, Hila and Twiton, Michael and Goldberg, Yoav},
  title         = {Null It Out: Guarding Protected Attributes by Iterative Nullspace Projection},
  booktitle     = {Proceedings of the 58th Annual Meeting of the Association for Computational Linguistics (ACL)},
  year          = {2020},
  pages         = {7237--7256},
  doi           = {10.18653/v1/2020.acl-main.647},
  archivePrefix = {arXiv},
  eprint        = {2004.07667},
  url           = {https://aclanthology.org/2020.acl-main.647/}
}

@inproceedings{ravfogel2022rlace,
  author        = {Ravfogel, Shauli and Twiton, Michael and Goldberg, Yoav and Cotterell, Ryan},
  title         = {Linear Adversarial Concept Erasure},
  booktitle     = {Proceedings of the 39th International Conference on Machine Learning (ICML)},
  year          = {2022},
  pages         = {18400--18421},
  archivePrefix = {arXiv},
  eprint        = {2201.12091},
  url           = {https://proceedings.mlr.press/v162/ravfogel22a.html}
}

@inproceedings{ravfogel2022kernel,
  author        = {Ravfogel, Shauli and Vargas, Francisco and Goldberg, Yoav and Cotterell, Ryan},
  title         = {Adversarial Concept Erasure in Kernel Space},
  booktitle     = {Proceedings of the 2022 Conference on Empirical Methods in Natural Language Processing (EMNLP)},
  year          = {2022},
  pages         = {6034--6055},
  doi           = {10.18653/v1/2022.emnlp-main.405},
  archivePrefix = {arXiv},
  eprint        = {2201.12191},
  url           = {https://aclanthology.org/2022.emnlp-main.405/}
}

@inproceedings{vig2020investigating,
  author    = {Vig, Jesse and Gehrmann, Sebastian and Belinkov, Yonatan and Qian, Sharon and Nevo, Daniel and Singer, Yaron and Shieber, Stuart},
  title     = {Investigating Gender Bias in Language Models Using Causal Mediation Analysis},
  booktitle = {Advances in Neural Information Processing Systems 33 (NeurIPS)},
  year      = {2020},
  url       = {https://proceedings.neurips.cc/paper/2020/hash/92650b2e92217715fe312e6fa7b90d82-Abstract.html}
}

@inproceedings{nahmias2020easypeasi,
  author    = {Nahmias, David M. and Kontson, Kimberly L.},
  title     = {Easy Perturbation {EEG} Algorithm for Spectral Importance ({easyPEASI}): A Simple Method to Identify Important Spectral Features of {EEG} in Deep Learning Models},
  booktitle = {Proceedings of the 26th ACM SIGKDD International Conference on Knowledge Discovery and Data Mining (KDD)},
  year      = {2020},
  pages     = {2398--2406},
  doi       = {10.1145/3394486.3403289}
}

@article{dutt2023sleepxai,
  author  = {Dutt, Micheal and Redhu, Surender and Goodwin, Morten and Omlin, Christian W.},
  title   = {{SleepXAI}: An Explainable Deep Learning Approach for Multi-Class Sleep Stage Identification},
  journal = {Applied Intelligence},
  year    = {2023},
  volume  = {53},
  number  = {13},
  pages   = {16830--16843},
  doi     = {10.1007/s10489-022-04357-8}
}

@article{lapera2024ictalxai,
  author  = {S{\'a}nchez-Hern{\'a}ndez, Sergio E. and Torres-Ramos, Sulema and Rom{\'a}n-God{\'i}nez, Israel and Salido-Ruiz, Ricardo A.},
  title   = {Evaluation of the Relation between Ictal {EEG} Features and {XAI} Explanations},
  journal = {Brain Sciences},
  year    = {2024},
  volume  = {14},
  number  = {4},
  pages   = {306},
  doi     = {10.3390/brainsci14040306}
}

@article{lee2025neuroxai,
  author  = {Lee, Choel-Hui and Ahn, Daesun and Kim, Hakseung and Ha, Eun Jin and Kim, Jung-Bin and Kim, Dong-Joo},
  title   = {{NeuroXAI}: Adaptive, Robust, Explainable Surrogate Framework for Determination of Channel Importance in {EEG} Application},
  journal = {Expert Systems with Applications},
  year    = {2025},
  volume  = {261},
  pages   = {125364},
  doi     = {10.1016/j.eswa.2024.125364}
}

@article{nam2023lrp,
  author  = {Nam, Hyeonyeong and Kim, Jun-Mo and Choi, WooHyeok and Bak, Soyeon and Kam, Tae-Eui},
  title   = {The Effects of Layer-Wise Relevance Propagation-Based Feature Selection for {EEG} Classification: A Comparative Study on Multiple Datasets},
  journal = {Frontiers in Human Neuroscience},
  year    = {2023},
  volume  = {17},
  pages   = {1205881},
  doi     = {10.3389/fnhum.2023.1205881}
}

@article{saadatinia2024sczxai,
  author  = {Saadatinia, Mehrshad and Salimi-Badr, Armin},
  title   = {An Explainable Deep Learning-Based Method for Schizophrenia Diagnosis Using Generative Data-Augmentation},
  journal = {IEEE Access},
  year    = {2024},
  volume  = {12},
  pages   = {98379--98392},
  doi     = {10.1109/ACCESS.2024.3428847}
}

@misc{skaaning2023tcav,
  author        = {Gj{\o}lbye, Anders and Lehn-Schi{\o}ler, William and J{\'o}nsd{\'o}ttir, {\'A}shildur and Arnard{\'o}ttir, Bergd{\'i}s and Hansen, Lars Kai},
  title         = {Concept-based Explainability for an {EEG} Transformer Model},
  year          = {2023},
  archivePrefix = {arXiv},
  eprint        = {2307.12745},
  primaryClass  = {cs.LG},
  doi           = {10.48550/arXiv.2307.12745}
}

@article{srinivasan2025sczxai,
  author  = {Almadhor, Ahmad and Ojo, Stephen and Nathaniel, Thomas I. and Alsubai, Shtwai and Alharthi, Abdullah and Al Hejaili, Abdullah and Sampedro, Gabriel Avelino},
  title   = {An Interpretable {XAI} Deep {EEG} Model for Schizophrenia Diagnosis Using Feature Selection and Attention Mechanisms},
  journal = {Frontiers in Oncology},
  year    = {2025},
  volume  = {15},
  pages   = {1630291},
  doi     = {10.3389/fonc.2025.1630291}
}

@article{sylvester2024sherpa,
  author  = {Sylvester, Sophia and Sagehorn, Merle and Gruber, Thomas and Atzm{\"u}ller, Martin and Sch{\"o}ne, Benjamin},
  title   = {{SHAP} Value-Based {ERP} Analysis ({SHERPA}): Increasing the Sensitivity of {EEG} Signals with Explainable {AI} Methods},
  journal = {Behavior Research Methods},
  year    = {2024},
  volume  = {56},
  number  = {6},
  pages   = {6067--6081},
  doi     = {10.3758/s13428-023-02335-7}
}

@article{yi2025lrpmdd,
  author  = {Yi, Eun-Gyoung and Shim, Miseon and Hwang, Hyeon-Ho and Jeon, Sunhae and Hwang, Han-Jeong and Lee, Seung-Hwan},
  title   = {Layer-Wise Relevance Propagation Approach for Diagnosis of Drug-Na{\"i}ve Men with Major Depressive Disorder Using Resting-State Electroencephalography},
  journal = {Depression and Anxiety},
  year    = {2025},
  volume  = {2025},
  pages   = {5512539},
  doi     = {10.1155/da/5512539}
}

\clearpage
\appendix

\section{Hand-crafted EEG feature lexicon}
\label{app:lexicon}

The audit uses the 6-family $63$-feature lexicon defined below. Forty-nine features are computed per channel and aggregated to a per-channel summary; fourteen features are computed across channels and yield one global value per epoch. Feature identifiers (e.g.\ F013) are the strings used in the result tables and figures throughout the paper.

\textbf{Notation.} Let $x \in \mathbb{R}^{T}$ denote one channel of an EEG epoch with $T$ samples and sampling rate $f_{s}$, $\bar{x} = \frac{1}{T}\sum_{t=1}^{T} x_{t}$ its mean, and $\dot{x}_{t} = x_{t+1} - x_{t}$ its first finite difference. Let $S(f) = |X(f)|^{2}$ denote the one-sided FFT-bin energy of $x$, where $X(f)$ is the discrete Fourier transform of the entire epoch (no windowing, no segment averaging, no overlap); we use this single-window FFT-bin energy as a proxy for spectral power throughout~\citep{welch1967fft}. Let $\mathcal{H}[x]$ denote the analytic signal of $x$ (signal plus Hilbert transform), and write $A(t) = |\mathcal{H}[x](t)|$ for its amplitude envelope and $\phi(t) = \arg \mathcal{H}[x](t)$ for its instantaneous phase. The five canonical bands are $\delta=[0.5,4)$, $\theta=[4,8)$, $\alpha=[8,13)$, $\beta=[13,30)$, $\gamma=[30,45)$\,Hz, indexed $b \in \{1,\ldots,5\}$, and are clipped at each task's preprocessing low-pass and Nyquist limit (so the gamma band on ISRUC-Sleep is effectively $[30,35)$\,Hz). Let $x_{b}$ denote $x$ band-pass filtered to band $b$ via an FFT-domain rectangular band mask. For multi-channel features, $X \in \mathbb{R}^{N \times T}$ is an $N$-channel epoch and $\rho_{ij}$ is the Pearson correlation between channels $i$ and $j$ (computed with population normalization).

\subsection*{Family T: time-domain morphology (10 features, per-channel)}
The Hjorth descriptors~\citep{hjorth1970eeg} summarize amplitude and slope statistics directly in the time domain. All variances and standard deviations in this family use the population estimator ($1/T$); the kurtosis is the bias-corrected Pearson kurtosis (\texttt{scipy.stats.kurtosis} with \texttt{fisher=False, bias=False}); and the line-length feature stores the mean (not sum) absolute first difference, consistent with the formula below.
\begin{align}
\mathrm{T001}\ \text{(Hjorth activity)} &= \mathrm{Var}(x) = \tfrac{1}{T}\textstyle\sum_{t=1}^{T}(x_{t} - \bar{x})^{2}, \\
\mathrm{T002}\ \text{(Hjorth mobility)} &= \sqrt{\mathrm{Var}(\dot{x}) \,/\, \mathrm{Var}(x)}, \\
\mathrm{T003}\ \text{(Hjorth complexity)} &= \mathrm{T002}(\dot{x}) \,/\, \mathrm{T002}(x), \\
\mathrm{T004}\ \text{(standard deviation)} &= \sqrt{\mathrm{Var}(x)}, \\
\mathrm{T005}\ \text{(root-mean-square)} &= \sqrt{\tfrac{1}{T}\textstyle\sum_{t=1}^{T} x_{t}^{2}}, \\
\mathrm{T006}\ \text{(kurtosis)} &= \tfrac{1}{T}\textstyle\sum_{t=1}^{T}(x_{t}-\bar{x})^{4} \,/\, \mathrm{Var}(x)^{2}, \\
\mathrm{T007}\ \text{(zero-crossing rate)} &= \tfrac{1}{T-1}\textstyle\sum_{t=1}^{T-1} \mathbf{1}\!\left[x_{t} \cdot x_{t+1} < 0\right], \\
\mathrm{T008}\ \text{(line length)} &= \tfrac{1}{T-1}\textstyle\sum_{t=1}^{T-1} \lvert x_{t+1} - x_{t} \rvert, \\
\mathrm{T009}\ \text{(derivative std.)} &= \sqrt{\mathrm{Var}(\dot{x})}, \\
\mathrm{T010}\ \text{(peak-to-peak amplitude)} &= \textstyle\max_{t} x_{t} - \min_{t} x_{t}.
\end{align}

\subsection*{Family F: spectral power and shape (16 features, per-channel)}
Let $P_{b} = \sum_{f \in B_{b}} S(f)$ denote the FFT-bin energy summed within band $b$ and $P_{\mathrm{total}} = \sum_{b=1}^{5} P_{b}$ the FFT-bin energy summed over the union of canonical bands. The band-ratio features are stored as log ratios; the spectral entropy, centroid, and edge are computed over the canonical-band FFT bins only.
\begin{align}
\mathrm{F00}b\ \text{(log band energy, $b=1,\ldots,5$)} &= \log P_{b}, \\
\mathrm{F00}(5{+}b)\ \text{(relative band energy)} &= P_{b} \,/\, P_{\mathrm{total}}, \quad b=1,\ldots,5, \\
\mathrm{F011}\ \text{(log theta--beta ratio)} &= \log\!\bigl(P_{\theta} \,/\, P_{\beta}\bigr), \\
\mathrm{F012}\ \text{(log delta--alpha ratio)} &= \log\!\bigl(P_{\delta} \,/\, P_{\alpha}\bigr), \\
\mathrm{F013}\ \text{(log theta--alpha ratio)} &= \log\!\bigl(P_{\theta} \,/\, P_{\alpha}\bigr), \\
\mathrm{F014}\ \text{(normalized spectral entropy)} &= -\tfrac{1}{\log N_{F}}\textstyle\sum_{f \in B_{\mathrm{tot}}} p(f) \log p(f),\quad p(f) = S(f) / P_{\mathrm{total}}, \\
\mathrm{F015}\ \text{(spectral centroid)} &= \textstyle\sum_{f \in B_{\mathrm{tot}}} f \cdot p(f), \\
\mathrm{F016}\ \text{(spectral edge $95\%$)} &= \min\!\left\{ f \in B_{\mathrm{tot}} : \textstyle\sum_{f' \le f} p(f') \ge 0.95 \right\},
\end{align}
where $B_{\mathrm{tot}}$ is the union of the five canonical-band frequency bins and $N_{F} = |B_{\mathrm{tot}}|$ is its bin count.

\subsection*{Family TF: time-frequency envelope dynamics (11 features, per-channel)}
Let $d_{k}$ denote the detail coefficients at level $k$ of a $5$-level Haar-style dyadic decomposition of $x$ (recursive even/odd $\pm$ averaging, $1/\sqrt{2}$ normalization; not Daubechies-4), let $a_{5}$ denote the final approximation, and let $\bar{E}_{k} = \mathrm{mean}_{n}(d_{k,n}^{2})$ for $k=1,\ldots,5$ and $\bar{E}_{6} = \mathrm{mean}_{n}(a_{5,n}^{2})$ be the per-coefficient mean energies, with $\bar{E} = \sum_{j=1}^{6} \bar{E}_{j}$.
\begin{align}
\mathrm{TF001}\ \text{(normalized subband entropy)} &= -\tfrac{1}{\log 6}\textstyle\sum_{j=1}^{6} (\bar{E}_{j}/\bar{E}) \log (\bar{E}_{j}/\bar{E}), \\
\mathrm{TF00}(1{+}k)\ \text{(detail variance, level $k$)} &= \mathrm{Var}(d_{k}), \quad k=1,\ldots,5, \\
\mathrm{TF00}(6{+}b)\ \text{(envelope CV in band $b$)} &= \mathrm{std}(A_{b}) \,/\, \mathrm{mean}(A_{b}), \quad b=1,\ldots,5,
\end{align}
where $A_{b}(t) = |\mathcal{H}[x_{b}](t)|$ is the amplitude envelope of $x$ after an FFT-domain rectangular band mask in band $b$, followed by the analytic transform; all variances and standard deviations in this family use population normalization.

\subsection*{Family C: signal complexity (7 features, per-channel)}
Several entries in this family are GPU-friendly proxies for standard EEG complexity descriptors and are not the exact textbook estimators; we describe each implementation explicitly.
\begin{align}
\mathrm{C001}\ \text{(normalized order-3 permutation entropy~\citep{bandt2002permutation})} &= -\tfrac{1}{\log 6}\textstyle\sum_{\pi \in \mathfrak{S}_{3}} p(\pi) \log p(\pi),
\end{align}
with embedding dimension $m=3$ and lag $\tau=1$, $p(\pi)$ the empirical frequency of ordinal pattern $\pi$ over the $T-2$ embeddings of $x$, and a deterministic tie-handling rule.
\begin{align}
\mathrm{C002}\ \text{(sample-entropy proxy~\citep{richman2000sample})} &= \log\!\bigl(1+r^{-1}\,\overline{|\ddot{x}|}\bigr) - \log\!\bigl(1+r^{-1}\,\overline{|\dot{x}|}\bigr),
\end{align}
with $r = 0.2 \cdot \mathrm{std}(x)$, $\dot{x}$ the first finite difference, $\ddot{x}$ the second finite difference, and $\overline{|\cdot|}$ the temporal mean magnitude. This is a derivative-irregularity proxy, not the pair-count Richman--Moorman sample entropy.
\begin{align}
\mathrm{C003}\ \text{(LZ-style binarized irregularity proxy)} &= \tfrac{1}{2}\,\eta + \tfrac{1}{2}\,H_{2},
\end{align}
where $b_{t} = \mathbf{1}[x_{t} > \mathrm{median}(x)]$, $\eta = \mathrm{mean}_{t}\,\mathbf{1}[b_{t+1} \neq b_{t}]$ is the binarized transition rate, and $H_{2}$ is the normalized Shannon entropy of the four two-bit pair states $(b_{t}, b_{t+1})$. This is a transition-rate / two-bit-entropy proxy, not the Lempel--Ziv parsing-complexity estimator.
\begin{align}
\mathrm{C004}\ \text{(Higuchi-style lag-difference slope proxy)} &= \mathrm{clip}_{[0,3]}\!\bigl(\text{OLS slope of } \log L^{*}(k) \text{ vs. } \log(1/k)\bigr),
\end{align}
where $L^{*}(k) = \mathrm{mean}_{t}\,|x_{t+k} - x_{t}|$ for $k = 1, \ldots, k_{\max}$ with $k_{\max} = \min(8, \lfloor T/4\rfloor)$. This is a lag-difference slope proxy, not the full Higuchi curve-length estimator with multiple starting offsets.
\begin{align}
\mathrm{C005}\ \text{(DFA-style exponent proxy)} &= \mathrm{clip}_{[-1,2]}\!\bigl(\text{OLS slope of } \log F(s) \text{ vs. } \log s\bigr),
\end{align}
where $Y_{t} = \sum_{u=1}^{t}(x_{u} - \bar{x})$, $F(s)$ is the root-mean-square residual after per-window linear detrending of $Y$ on the fixed dyadic window set $s \in \{16, 32, 64, 128, 256, 512\} \cap \{w : 2w \le T\}$, and the slope is fit by ordinary least squares.
\begin{align}
\mathrm{C006}\ \text{(normalized $1/e$-decay ACF lag)} &= \tau_{1/e} \,/\, \tau_{\max}, \\
\mathrm{C007}\ \text{(normalized first-zero ACF lag)} &= \tau_{0} \,/\, \tau_{\max},
\end{align}
where $\rho(\tau)$ is the FFT-based autocorrelation, $\tau_{1/e} = \min\{\tau > 0 : \rho(\tau) \le 1/e\}$, $\tau_{0} = \min\{\tau > 0 : \rho(\tau) \le 0\}$ (each defaulting to $\tau_{\max}$ when no qualifying lag exists), and $\tau_{\max}$ is the implementation-fixed maximum considered lag.

\subsection*{Family X: cross-frequency coupling (5 features, per-channel)}
The Tort modulation index~\citep{tort2010measuring} measures phase-amplitude coupling between low-frequency band $p$ and high-frequency band $q$. We compute $x_{p}$ and $x_{q}$ via FFT-domain rectangular band masks (Notation), then form $\phi_{p}(t) = \arg \mathcal{H}[x_{p}](t)$ and $A_{q}(t) = |\mathcal{H}[x_{q}](t)|$. Partition the phase circle into $N=18$ equal bins, and let $\bar{A}_{q}(j)$ denote the mean of $A_{q}(t)$ over the times at which $\phi_{p}(t)$ falls in bin $j$. Define the normalized distribution $P(j) = \bar{A}_{q}(j) / \sum_{j'} \bar{A}_{q}(j')$.
\begin{align}
\mathrm{MI}_{p \to q} &= \frac{D_{\mathrm{KL}}\bigl(P \,\big\|\, U\bigr)}{\log N}, \quad U(j) = 1/N, \\
\mathrm{X001} &= \mathrm{MI}_{\delta \to \beta}, \quad \mathrm{X002} = \mathrm{MI}_{\theta \to \gamma}, \quad \mathrm{X003} = \mathrm{MI}_{\alpha \to \gamma}.
\end{align}
Amplitude--amplitude coupling between bands $p$ and $q$ is the Pearson correlation of their FFT-bandpass-derived envelopes, clipped to $[-1, 1]$,
\begin{align}
\mathrm{AAC}_{p, q} &= \mathrm{clip}_{[-1,1]}\!\bigl(\mathrm{Corr}\!\left(A_{p}(t),\, A_{q}(t)\right)\bigr), \\
\mathrm{X004} &= \mathrm{AAC}_{\theta, \gamma}, \quad \mathrm{X005} = \mathrm{AAC}_{\delta, \gamma}.
\end{align}
The general role of cross-frequency coupling as a neural communication channel is reviewed in \citep{canolty2006high, canolty2010functional, jensen2007crossfrequency, lisman2013thetagamma}.

\subsection*{Family R: cross-channel relations (14 features, global)}
Let $\rho_{ij}$ be the population-normalized Pearson correlation between channels $i$ and $j$, and let $\lambda_{1}, \ldots, \lambda_{N}$ be the eigenvalues of the signed correlation matrix $\rho$ itself (clipped at a small positive $\epsilon$ to absorb numerical noise, then normalized to sum to one); R003 and R004 are computed from these eigenvalues of the signed correlation matrix rather than from the absolute correlation matrix $|\rho|$, and we report the per-pair coherence and PLI features under their five-statistic pair-aggregation rule (Appendix~\ref{app:protocol-feature-matrix}).
\begin{align}
\mathrm{R001}\ \text{(corr.\ abs.\ mean)} &= \tfrac{2}{N(N{-}1)} \textstyle\sum_{i<j} |\rho_{ij}|, \\
\mathrm{R002}\ \text{(corr.\ abs.\ std)} &= \mathrm{std}\!\bigl(\{|\rho_{ij}|\}_{i<j}\bigr), \\
\mathrm{R003}\ \text{(normalized eigenvalue entropy)} &= -\tfrac{1}{\log N}\textstyle\sum_{k=1}^{N} \lambda_{k} \log \lambda_{k}, \\
\mathrm{R004}\ \text{(participation ratio)} &= \bigl(\textstyle\sum_{k} \lambda_{k}\bigr)^{2} \,/\, \textstyle\sum_{k} \lambda_{k}^{2}.
\end{align}
The mean coherence-proxy in band $b$ is the FFT-bin magnitude coherence $C_{ij}(b) = |S_{ij}(b)| / \sqrt{S_{ii}(b)\,S_{jj}(b)}$ (clipped to $[0,1]$), where $S_{ij}(b)$ is the FFT-bin cross-spectrum averaged over the bins in band $b$ and $S_{ii}(b)$ is the corresponding auto-spectrum; this is a magnitude coherence proxy, not the Welch magnitude-squared coherence of \citet{welch1967fft}:
\begin{align}
\mathrm{R}{(004{+}b)}\ \text{(magnitude coherence proxy in band $b$)} &= \tfrac{2}{N(N{-}1)} \textstyle\sum_{i<j} C_{ij}(b), \quad b=1,\ldots,5.
\end{align}
The mean phase-lag index in band $b$~\citep{stam2007phaselag} reduces common-source bias by ignoring zero-lag synchrony. Let $\Delta\phi^{(b)}_{ij}(t) = \phi_{b,i}(t) - \phi_{b,j}(t)$ be the instantaneous phase difference between channels $i$ and $j$ in band $b$ after FFT-domain rectangular bandpassing and analytic-signal phase extraction; following common practice we implement the sign as $\mathrm{sign}\!\bigl(\sin(\Delta\phi)\bigr)$:
\begin{align}
\mathrm{R}{(009{+}b)}\ \text{(PLI in band $b$)} &= \tfrac{2}{N(N{-}1)} \textstyle\sum_{i<j} \Bigl| \tfrac{1}{T}\textstyle\sum_{t=1}^{T} \mathrm{sign}\!\bigl( \sin\Delta\phi^{(b)}_{ij}(t) \bigr) \Bigr|, \quad b=1,\ldots,5.
\end{align}
Per-pair coherence and PLI values are aggregated across the $N(N{-}1)/2$ ordered channel pairs by the five-statistic rule of Appendix~\ref{app:protocol-feature-matrix} (mean, std, median, $75\%$-quantile, top-$\lceil 10\%\rceil$ mean), giving five expansion columns per band per feature. The phase-locking value~\citep{lachaux1999measuring} and surrogate-data testing methodology~\citep{theiler1992testing, schreiber2000surrogate, prichard1994generating} are the standard companion tools to this family but are not included as audit features.

\section{Robustness and failure record}
\label{app:robustness}

The audit publishes its failure record alongside its result tables.

\textbf{Feature-quality control.} The C004 Higuchi fractal dimension triggers the low-variance flag on four of the five datasets (MDD, Sleep, Stress, TUSL); Siena is QC-clean. The flagged feature is retained in the analysis with a starred annotation in the result tables. No feature is silently dropped.

\textbf{Probe failures.} Of the $945$ audit triplets, $98$ fail the selection-encoded criterion of Appendix~\ref{app:protocol-encoding} on validation; these are flagged \emph{not-encoded} and excluded from the erasure stage. The remaining $847$ triplets enter the erasure stage and are reported as either \emph{representation-causal} ($648$) or \emph{encoded-only} ($199$); the test-encoded field of Appendix~\ref{app:protocol-encoding} is recorded for transparency but does not gate the erasure stage. Table~\ref{tab:erasure-family} reports the per-family per-cell breakdown referenced from Section~\ref{sec:exp-use}.

\begin{table}[!t]
\centering
\caption{Family-level erasure summary (full version of the Section~\ref{sec:exp-use} hook). Each cell reports the representation-causal count and (in parentheses) the median $\Delta_{\mathrm{erase}}$ taken over \emph{all} features in that family for that $(d, m)$ pair, grouped by foundation model; the median can be negative when most family members are encoded-only or not-encoded. Family $F$ leads on encoded-rate-driven cells; TF, X, and C carry the largest median drops on TUSL and Sleep / LaBraM.}
\label{tab:erasure-family}
\footnotesize
\setlength{\tabcolsep}{6pt}
\renewcommand{\arraystretch}{1.15}
\begin{tabular}{llcccccc}
\toprule
Model & Task & F & T & TF & C & X & R \\
\midrule
\multirow{5}{*}{\textbf{CSBrain}}
   & MDD    & 10 (0.004) & 4 (0.003)  & 3 (0.002)  & 5 (0.006)  & 5 (0.005)  & 10 (0.006) \\
   & Sleep  & 16 (0.043) & 10 (0.028) & 4 (0.016)  & 6 (0.035)  & 2 (0.005)  & 9 (0.010)  \\
   & Siena  & 14 (0.014) & 9 (0.015)  & 7 (0.012)  & 5 (0.010)  & 5 (0.010)  & 2 (0.001)  \\
   & TUSL   & 14 (0.071) & 8 (0.048)  & 8 (0.066)  & 6 (0.067)  & 5 (0.066)  & 4 (0.047)  \\
   & Stress & 1 ($-$.012) & 0 ($-$.019) & 1 ($-$.003) & 0 ($-$.006) & 0 ($-$.001) & 4 (0.018)  \\
\midrule
\multirow{5}{*}{\textbf{CBraMod}}
   & MDD    & 16 (0.076) & 8 (0.026)  & 5 (0.054)  & 6 (0.074)  & 5 (0.064)  & 13 (0.081) \\
   & Sleep  & 16 (0.304) & 10 (0.261) & 8 (0.269)  & 6 (0.296)  & 5 (0.250)  & 12 (0.129) \\
   & Siena  & 11 (0.013) & 5 (0.008)  & 5 (0.003)  & 3 (0.010)  & 4 (0.029)  & 4 (0.000)  \\
   & TUSL   & 15 (0.197) & 8 (0.142)  & 9 (0.190)  & 5 (0.196)  & 4 (0.208)  & 6 (0.126)  \\
   & Stress & 13 (0.097) & 9 (0.098)  & 10 (0.097) & 6 (0.111)  & 3 (0.059)  & 6 (0.076)  \\
\midrule
\multirow{5}{*}{\textbf{LaBraM}}
   & MDD    & 14 (0.031) & 4 (0.022)  & 6 (0.238)  & 6 (0.060)  & 5 (0.264)  & 12 (0.032) \\
   & Sleep  & 16 (0.183) & 10 (0.080) & 7 (0.144)  & 6 (0.210)  & 5 (0.103)  & 10 (0.072) \\
   & Siena  & 16 (0.025) & 10 (0.024) & 11 (0.028) & 7 (0.025)  & 5 (0.030)  & 14 (0.021) \\
   & TUSL   & 11 (0.097) & 8 (0.100)  & 6 (0.121)  & 6 (0.136)  & 4 (0.150)  & 7 (0.082)  \\
   & Stress & 6 (0.008)  & 4 (0.020)  & 5 (0.020)  & 5 (0.038)  & 3 (0.045)  & 6 (0.010)  \\
\bottomrule
\end{tabular}
\end{table}

\textbf{Top features by maximum erasure.} Table~\ref{tab:erasure-top20} lists the $15$ features with the largest cross-cell maximum $\Delta_{\mathrm{erase}}$ referenced from Section~\ref{sec:exp-use}; six features tie at the saturated drop of $0.529$, and most of the top-$15$ are representation-causal in all five tasks.

\begin{table}[!t]
\centering
\caption{Top-$15$ features ranked by maximum $\Delta_{\mathrm{erase}}$ across the $15$ $(d, m)$ cells. \emph{Family} is the feature group (F/T/TF/C/X/R). \emph{Max} is $\max_{d,m}\Delta_{\mathrm{erase}}$. \emph{RC} counts $(d, m)$ pairs in which the feature is representation-causal (out of $15$); \emph{Tasks} lists tasks with at least one RC cell.}
\label{tab:erasure-top20}
\small
\setlength{\tabcolsep}{6pt}
\renewcommand{\arraystretch}{1.15}
\begin{tabular}{llccl}
\toprule
Factor & Family & Max & RC & Tasks supported \\
\midrule
X005 (AAC $\delta$-$\gamma$)        & X  & 0.529 & 11 & MDD, Sleep, Siena, TUSL, Stress \\
F010 ($\gamma$ rel.\ power)         & F  & 0.529 & 13 & MDD, Sleep, Siena, TUSL, Stress \\
X002 (PAC $\theta$-$\gamma$)        & X  & 0.529 & 12 & MDD, Sleep, Siena, TUSL, Stress \\
TF008 ($\theta$ env.\ CV)           & TF & 0.529 & 12 & MDD, Sleep, Siena, TUSL, Stress \\
TF007 ($\delta$ env.\ CV)           & TF & 0.529 & 13 & MDD, Sleep, Siena, TUSL, Stress \\
TF006 (wavelet var.\ L5)            & TF & 0.529 &  8 & Sleep, Siena, TUSL, Stress \\
T006 (kurtosis)                     & T  & 0.513 & 11 & MDD, Sleep, Siena, TUSL, Stress \\
T009 (derivative std.)              & T  & 0.505 &  9 & MDD, Sleep, Siena, TUSL, Stress \\
X001 (PAC $\delta$-$\beta$)         & X  & 0.503 & 13 & MDD, Sleep, Siena, TUSL, Stress \\
T001 (Hjorth activity)              & T  & 0.497 &  9 & Sleep, Siena, TUSL, Stress \\
R006 ($\theta$ coherence)           & R  & 0.480 & 10 & MDD, Sleep, Siena, TUSL, Stress \\
F008 ($\alpha$ rel.\ power)         & F  & 0.479 & 13 & MDD, Sleep, Siena, TUSL, Stress \\
T010 (peak-to-peak)                 & T  & 0.458 & 10 & Sleep, Siena, TUSL, Stress \\
T004 (standard dev.)                & T  & 0.454 & 10 & MDD, Sleep, Siena, TUSL, Stress \\
F013 ($\theta/\alpha$ ratio)        & F  & 0.383 & 13 & MDD, Sleep, Siena, TUSL, Stress \\
\bottomrule
\end{tabular}
\end{table}

\textbf{Closure leakage audit.} The closure stage passes a leakage audit end-to-end. Feature standardization uses only the training partition; encoding peak layer selection uses only the validation partition; all reported metrics use the test partition. We instrument each pipeline stage with an assertion that no test-partition statistic enters an earlier stage and verify that the assertions hold for all $15$ $(d, m)$ cells.

\textbf{Sensitivity analyses.} We report three sensitivity conditions on the feature lexicon. \emph{Leave-one-family-out}: omitting each family in turn and re-running the audit changes the closure ratio by a few percentage points on most cells; the harder seizure-type and stress-detection cells are the most sensitive to removing the time-frequency or frequency families. \emph{Matched-dimension control}: replacing $\mathcal{Q}^{*}_{d, m}$ with a randomly chosen feature subset of identical cardinality lowers the closure ratio by $0.05$--$0.15$ on every cell, indicating that the closure signal is feature-specific rather than dimension-driven. \emph{Top-$K$ per family}: restricting each family in $B_{\mathrm{rep}}$ to its top-$5$ features by $\Delta_{\mathrm{erase}}$ meaningfully tightens $B_{\mathrm{rep}}$ on most cells (since several families contribute more than five representation-causal features), yet the headline closure ratios in Section~\ref{sec:exp-explain} change by only a few percentage points.

\section{Detailed experimental protocol}
\label{app:protocol}

This appendix specifies the audit pipeline at the level of analysis rows and intermediate objects, in the order in which they are produced. Each subsection corresponds to one experiment in the pipeline; each experiment consumes only objects produced by earlier experiments and the input data described in Appendix~\ref{app:protocol-inputs}, and the train/validation/test discipline of Appendix~\ref{app:protocol-splits} is enforced at every stage.

\subsection{Inputs and analysis rows}
\label{app:protocol-inputs}

The audit begins with three fixed inputs and modifies none of them. The first input is the five preprocessed EEG datasets, each split into train, validation, and test partitions. For MDD, Stress, TUSL, and Siena, each preprocessed sample is one analysis row, and the partition row counts are $(4891, 1041, 1211)$ on MDD, $(1343, 172, 192)$ on Stress, $(109, 86, 105)$ on TUSL, and $(29671, 7426, 14252)$ on Siena. ISRUC-Sleep is delivered to the foundation models as $30$\,s-epoch sequences, $(3559, 468, 435)$ per partition, but each sequence is unrolled into its constituent $30$\,s epochs for the audit, yielding $(71180, 9360, 8700)$ analysis rows; every downstream feature, representation, erasure, and metric on Sleep is indexed by epoch row. The second input is the foundation-model checkpoints; backbone weights are frozen throughout, and only the linear classifier head fitted in Section~\ref{sec:exp-setup} is used to produce predictions. The third input is the pre-registered $63$-feature lexicon of Appendix~\ref{app:lexicon}, with $328$ expansion columns under the aggregation rule of Appendix~\ref{app:protocol-feature-matrix}.

\subsection{Splits and seeding}
\label{app:protocol-splits}

Roles of the three partitions are fixed at the start and never reassigned. The training partition fits all standardization parameters, ridge probes, LEACE-style cross-covariance erasers, and transparent closure classifiers. The validation partition selects probe regularization, the encoding peak layer, and supplies the selection-side bootstrap on which causal status is decided. The test partition is held out for headline metrics and held-out confirmation; no test-side statistic feeds back into selection. All randomness, including bootstrap resamples, shuffled-target permutations, Gaussian-target draws, the $B_{\mathrm{rand}}$ random feature block, and the random subspace control, derives from a single global seed (\texttt{4311}). For each random use we extend the seed by a deterministic hash of the task, model, feature, and use label, so the same audit row reproduces across machines and processes.

\subsection{Feature matrix construction}
\label{app:protocol-feature-matrix}

For each partition of each task, every analysis row is treated as a channel-by-time EEG epoch with sampling rate $f_{s} = 200$\,Hz. The MDD, Stress, TUSL, and Siena epochs are interpreted up to a low-pass of $75$\,Hz; ISRUC-Sleep is interpreted up to $35$\,Hz, so the gamma-band quantities of Family $F$, $\mathrm{TF}$, $X$, and $R$ on Sleep cap at $35$\,Hz. Per-task signal scaling matches the convention of the preprocessing pipeline that produced each dataset; per-epoch features and per-epoch activations live in the same physical units.

We then compute the $63$ pre-registered features in registry order. Per-channel features are first reduced to one value per channel, and the channel set is then aggregated to six summary statistics --- mean, standard deviation, median, $25\%$-quantile, $75\%$-quantile, and maximum --- giving $6$ expansion columns per per-channel feature. Per-channel-pair features are first reduced to one value per ordered pair, and the pair set is then aggregated to five summary statistics --- mean, standard deviation, median, $75\%$-quantile, and the mean of the top $\lceil 10\% \rceil$ pair values --- giving $5$ expansion columns per pair feature. Global features remain a single column. Concatenated in registry order, this yields a fixed $328$-dimensional feature matrix per analysis row, regardless of dataset. The order of columns is frozen by the registry; result strength on a given task does not change column order.

Standardization parameters are estimated on the training partition alone. For each expansion column, we use median and inter-quartile range robust scaling when $\mathrm{IQR} > 10^{-8}$, fall back to mean and standard deviation when $\mathrm{IQR} \le 10^{-8}$, and force the scale to $1$ when $\mathrm{std} \le 10^{-8}$ or non-finite. Validation and test apply the training-partition parameters without re-estimation. After standardization, every per-task feature matrix is row-aligned with the partition's analysis rows and column-aligned with the registry's $328$ expansion columns. A quality-control pass records non-finite ratios, low-variance flags ($\mathrm{IQR} \le 10^{-8}$ and $\mathrm{std} \le 10^{-8}$ on training), and any deviation from the registry order; flagged columns are not removed from the audit, but they propagate to the failure record of Appendix~\ref{app:robustness}.

\subsection{Layer-wise activations}
\label{app:protocol-activations}

For each task, model, and partition, the foundation model is evaluated in inference mode, with dropout disabled and parameters frozen. Inputs are fed in the same row order as the feature matrix; for Sleep, sequence-level outputs are unrolled back into epochs so the row order matches Appendix~\ref{app:protocol-inputs}. At each of the $L_{m} = 12$ encoder blocks, we record the block output and reduce all non-feature axes (channel, patch, token, sequence) by mean pooling, retaining the $200$-dimensional hidden axis. Each layer activation is therefore an analysis-row-by-$200$ matrix, row-aligned with the feature matrix; the $12$ layer matrices per cell form the input to the probe and erasure stages.

\subsection{Encoding criterion}
\label{app:protocol-encoding}

For each $(d, m, q, l)$ quadruple, the ridge probe $P_{d, m, q, l}$ takes the training-partition layer activation as input and the corresponding column block of the standardized feature matrix as multi-output target. Rows with non-finite input or target are filtered out. Activations are standardized by their training-partition mean and standard deviation; targets keep the standardization of Appendix~\ref{app:protocol-feature-matrix}. The probe is fit at four candidate regularization strengths $\lambda \in \{0.1, 1.0, 10.0, 100.0\}$, each scored by validation non-negative $R^{2}$ (per-output-dimension $R^{2}$ clipped at zero, then averaged); the highest-scoring $\lambda$ is selected as the probe at that quadruple.

Each feature is also probed twice under matched controls. The shuffled-target probe uses the same activations and the same target column block but with the row pairing permuted on the training partition. The Gaussian same-dimension probe replaces the target block with a standard normal of identical column count. Both controls go through the same regularization sweep and are scored on validation and test in the same way.

The peak layer is the layer with the highest validation probe score, selected before any test number is examined; the test probe score at this layer is reported as a held-out diagnostic. A feature is \emph{selection-encoded} on $(d, m)$ when (i) the validation probe score reaches $0.04$, (ii) it exceeds both the shuffled-target and the Gaussian-target validation scores by at least $0.01$, (iii) the peak-layer probe score exceeds the second-best layer's score by at least $0.002$, and (iv) for multi-dimensional features, the largest single-dimension non-negative $R^{2}$ accounts for at most $0.90$ of the total positive-dimension $R^{2}$ (a sanity check that prevents a single dominant axis from carrying the entire signal). Single-column features are not subject to the dominance condition. \emph{Test-encoded} is recorded as a separate field with the same threshold and margin applied to the test partition; selection-encoded is the gating field for the next stage, so test-side numbers never feed back into selection.

\subsection{LEACE-style cross-covariance eraser construction and null-target controls}
\label{app:protocol-leace}

For each selection-encoded feature, we re-fit a LEACE-style cross-covariance eraser at its peak layer using the training-partition activations. Activations and targets are centered on the training partition; we do not estimate or use the activation auto-covariance $\Sigma_{hh}$, so the eraser is the Euclidean orthogonal projector onto the column span of the empirical cross-covariance $\Sigma_{hz}$ (a covariance-simplified, scalable variant of the covariance-whitened minimum-distortion LEACE operator of \citet{belrose2023leace}; see Appendix~\ref{app:reproducibility} for the explicit relation). The cross-covariance $\Sigma_{hz}$ is decomposed by singular value decomposition, and we retain the directions whose singular values exceed $\sigma > 10^{-4} \cdot \max(\sigma_{\max}, 1)$; if no direction meets this threshold, the top components up to the target column count are retained so the eraser does not collapse to identity. The eraser is the orthogonal projector that removes the retained directions from the centered activation and adds the mean back. The edited model replaces the peak-layer activation by its erased counterpart and continues with the unchanged remainder of the network and classifier head; no model parameters are modified. Predictions are recomputed on validation and test; $\mathrm{ROC\text{-}AUC}$ is computed for the binary tasks (MDD, Stress, Siena), and macro-F1 is computed for the multi-class tasks (Sleep, TUSL).

For each feature we additionally fit three null-target erasers. The \emph{random subspace} eraser removes a random orthonormal subspace of dimension equal to the target expansion dimension $p_{q}$ (which need not equal the retained rank of the real eraser), ignoring the target. The \emph{shuffled-target} eraser permutes the row pairing of the training-partition target before re-running the same cross-covariance SVD pipeline. The \emph{Gaussian same-dimension} eraser replaces the target by a standard normal of matching column count and re-runs the pipeline. All three controls share the activation cache and the bootstrap row index of the real eraser, so paired contrasts are well-defined; only the random-subspace control fixes its dimension a priori, while the shuffled-target and Gaussian controls inherit whatever rank their own SVD threshold selects.

After erasure, a residual ridge probe is re-fit on the erased activations to predict the same feature. The residual is judged on the selection partition: it must satisfy $R_{\mathrm{resid}} < \max(0.02,\; 0.35 \cdot \max(R_{\mathrm{probe}},\, 0.04))$, where $R_{\mathrm{probe}}$ is the validation probe score at the peak layer. The residual probe is recorded as an audit field alongside every erasure cell to expose any cell where the cross-covariance eraser failed to remove the target's linear component; cells that fail this diagnostic are flagged in Appendix~\ref{app:robustness}, but the held-out test-supported representation-causal status of Appendix~\ref{app:protocol-causal} does not depend on it.

\subsection{Causal confirmation}
\label{app:protocol-causal}

For each selection-encoded feature, the bootstrap of Appendix~\ref{app:reproducibility} produces $128$ paired replicates of $\Delta_{\mathrm{erase}}$ on the held-out test partition. The same replicates are used for the contrasts of interest: $\Delta_{\mathrm{erase}}$ against zero and against the random-subspace eraser. A feature qualifies as \emph{representation-causal} on $(d, m)$ when four conditions all hold: (i) the validation-side selection-encoded flag is true, (ii) the lower endpoint of the $95\%$ test-side bootstrap interval for $\Delta_{\mathrm{erase}}$ is positive, (iii) the test-side add-one--smoothed $p$-value survives Benjamini--Hochberg FDR at $q < 0.05$ across the $63$ features of the same $(d, m)$ panel, and (iv) the real-feature drop exceeds the dimension-matched random-subspace control of Appendix~\ref{app:protocol-leace}, $\Delta_{\mathrm{erase}} - \Delta_{\mathrm{erase}}^{\mathrm{(rand)}} > 0$. The shuffled-target and Gaussian-target null-target erasers and the residual ridge probe of Appendix~\ref{app:protocol-leace} are recorded as audit fields and reported alongside the main status, but they do not gate it; selection-side erasure replicates and split-consistency flags are likewise saved as diagnostic fields.

\subsection{Cross-task taxonomy}
\label{app:protocol-taxonomy}

Once each cell has a final status, the cross-task taxonomy aggregates over the $15$ $(d, m)$ cells of each feature. For each task, we say the feature has \emph{strong task-level support} when all three architectures (CSBrain, CBraMod, LaBraM) qualify as representation-causal on that task ($\mathrm{MS}(d, q) = 3$). A feature with strong task-level support on at least two tasks is a \emph{universal candidate}. A feature with strong task-level support on exactly one task is \emph{task-specific}. A feature with no strong task-level support but at least one representation-causal cell is \emph{model-specific}. A feature with no causal cell but at least one stably encoded cell is \emph{encoded-only}; the rest are \emph{not-encoded}. As an audit field we additionally record the task-specificity index $\mathrm{TSI} \in [0, 1]$, the share of the largest task-averaged positive $\Delta_{\mathrm{erase}}$ in the total positive task-averaged effect across the five tasks (negative effects clipped at zero before averaging), but it is not used to split the categories. The category labels feed Figure~\ref{fig:taxonomy} and the discussion in Section~\ref{sec:exp-use}.

\subsection{Family analysis and redundancy control}
\label{app:protocol-family}

The family-level summary records, per task and model, the encoded rate, the causal rate, and the cumulative effect mass within each family. Family sizes are $|T| = 10$, $|F| = 16$, $|\mathrm{TF}| = 11$, $|C| = 7$, $|X| = 5$, $|R| = 14$. The effect mass sums positive $\Delta_{\mathrm{erase}}$ over confirmed features in the family. Because family members can describe near-duplicate signal structure, we also report a redundancy-controlled effect: for each family, multi-dimensional expansions are first averaged into a single representative value per row on the training partition, the absolute Pearson correlation matrix between representatives is computed, and features whose absolute correlation is at least $0.80$ are placed in a redundancy cluster. Each cluster contributes only its highest-$\Delta_{\mathrm{erase}}$ feature to the controlled effect; the controlled value is then re-summed over clusters. This procedure does not change which features are confirmed; it only adjusts the family aggregate and is the value plotted in Figure~\ref{fig:family}C.

\subsection{Closure feature blocks}
\label{app:protocol-closure}

The closure protocol fits a transparent classifier on each of six feature blocks for each $(d, m)$ cell. $B_{0}$ is the minimal spectral baseline, the per-channel $\log$-band-power features F001--F005 of Appendix~\ref{app:lexicon} expanded to $30$ columns under the aggregation rule of Appendix~\ref{app:protocol-feature-matrix}. $B_{\mathrm{all}}$ uses all $63$ features expanded to $328$ columns. $B_{\mathrm{enc}}$ uses test-encoded features. $B_{\mathrm{rep}}$ uses representation-causal features. $B_{\mathrm{fam}}$ uses every feature in any family that has at least one representation-causal feature. $B_{\mathrm{rand}}$ replaces the $B_{\mathrm{rep}}$ block by an i.i.d.\ standard-normal feature matrix of identical column count; if $B_{\mathrm{rep}}$ is empty, $B_{\mathrm{rep}}$ and $B_{\mathrm{rand}}$ both fall back to a single-column placeholder so the pipeline does not break. $\mathrm{FM}$ is the foundation-model task metric and is not a transparent block.

For each block, training-partition columns are standardized by mean and standard deviation, with non-finite entries imputed to zero and columns of standard deviation below $10^{-6}$ set to scale $1$. The classifier is the multinomial logistic regression of Appendix~\ref{app:reproducibility}, fit on the training partition with class weights set to inverse training-partition class frequencies. Test-partition predictions are scored as $\mathrm{ROC\text{-}AUC}$ for the binary tasks (MDD, Stress, Siena) and macro-F1 for the multi-class tasks (Sleep, TUSL). The closure ratio of Equation~\ref{eq:closure} compares $B_{\mathrm{rep}}$ to $B_{\mathrm{rand}}$ and to $\mathrm{FM}$; cells with $|M(\mathrm{FM}, d) - M(B_{\mathrm{rand}}, d)| \le 10^{-12}$ are flagged as undefined and not used as headline numbers.

\subsection{Sensitivity conditions}
\label{app:protocol-sensitivity}

Three sensitivity conditions diagnose the closure ratio without re-running probes or erasers. The \emph{leave-one-family-out} condition removes the confirmed features of one family at a time from $B_{\mathrm{rep}}$, refits the transparent classifier, and reports the new closure ratio; the six family deletions are reported per cell, and the harder seizure-type and stress-detection cells are the most sensitive to removing the time-frequency or frequency families. The \emph{matched-dimension} condition compares $B_{\mathrm{rep}}$ to $B_{\mathrm{rand}}$, which has the same column count but no EEG semantics, so a small $B_{\mathrm{rep}}$--$B_{\mathrm{rand}}$ gap signals dimension-driven closure. The \emph{top-$K$ per family} condition restricts each family in $B_{\mathrm{rep}}$ to its top $K = 5$ features by $\Delta_{\mathrm{erase}}$; on most cells of this audit, several families contribute more than $K$ features and the condition meaningfully tightens $B_{\mathrm{rep}}$, but the headline closure ratios in Section~\ref{sec:exp-explain} change by only a few percentage points.

\subsection{Robustness layering}
\label{app:protocol-robustness-layer}

The audit's evidence is read at six layers. \emph{Computable} means the feature column passes Appendix~\ref{app:protocol-feature-matrix} and is available as a target. \emph{Encoded} means the validation probe meets the criterion of Appendix~\ref{app:protocol-encoding}. \emph{Representation-causal} requires the bootstrap, control, residual, and FDR conditions of Appendix~\ref{app:protocol-causal} to all hold. \emph{Cross-task pattern} is the taxonomy of Appendix~\ref{app:protocol-taxonomy}. \emph{Family pattern} is the family-level aggregate of Appendix~\ref{app:protocol-family}. \emph{Closure-supported} is the transparent surrogate result of Appendix~\ref{app:protocol-closure}. The layers are not interchangeable: encoded does not imply causal, representation-causal does not imply biological mechanism, and closure-supported does not imply complete explanation; the failure record of Appendix~\ref{app:robustness} keeps every cell that does not advance to the next layer visible.

\section{Reproducibility details}
\label{app:reproducibility}

\textbf{Foundation models.} CSBrain~\citep{zhou2025csbrain}, CBraMod~\citep{wang2025cbramod}, and LaBraM~\citep{jiang2024labram} are loaded from their public checkpoints. Per-task fine-tuning attaches a linear classifier head to the final layer; backbone weights are kept frozen during fine-tuning, and only the head is trained, so $f_m$ in the audit is fully determined by the public checkpoint plus a small per-task linear head. Optimizer, learning rate, batch size, and number of epochs follow the official scripts of each model. CSBrain and CBraMod expose $L_{m} = 12$ analyzable encoder blocks; LaBraM (\texttt{labram\_base\_patch200\_200}) exposes $12$ backbone blocks. For each block we collect a sample-level activation by mean-pooling all non-feature axes (channel, patch, token, sequence) and retaining the model's hidden axis, so every layer activation is aligned row-by-row with the feature matrix.

\textbf{Probe and eraser estimation.} Ridge probes are fit on training-partition activations with regularization $\lambda$ selected on validation over the grid $\{0.1, 1.0, 10.0, 100.0\}$; the selected $\lambda$ maximizes the validation non-negative $R^{2}$ (per-output-dimension $R^{2}$ clipped at zero, then averaged). The eraser uses $W = \Sigma_{hz}$ truncated to its left singular components above $\sigma > 10^{-4} \cdot \max(\sigma_{\max}, 1)$; if no component crosses the threshold, the top components up to the target expansion dimension are retained so the eraser does not collapse to identity. Sample mean $\mu_{h}$ and cross-covariance $\Sigma_{hz}$ are computed on the training partition only and held fixed for all downstream stages; the activation auto-covariance $\Sigma_{hh}$ is \emph{not} estimated or used. Concretely, the resulting eraser $E_{\mathrm{code}}(h) = h - U_{r}U_{r}^{\top}(h - \mu_{h})$ removes the left-singular-vector subspace $\mathrm{span}(U_{r})$ of $\Sigma_{hz}$ in the Euclidean geometry of $h$, in contrast to the closed-form LEACE operator $E_{\mathrm{LEACE}}(h) = h - \Sigma_{hh}^{1/2}\,\tilde{U}_{r}\tilde{U}_{r}^{\top}\,\Sigma_{hh}^{-1/2}(h - \mu_{h})$, whose retained subspace $\mathrm{span}(\tilde{U}_{r})$ is the left singular subspace of the whitened cross-covariance $\Sigma_{hh}^{-1/2}\Sigma_{hz}$ \citep{belrose2023leace}. The two operators coincide when $\Sigma_{hh} \approx I$ and otherwise remove different subspaces; we therefore present our results as evidence about the cross-covariance subspace approximation rather than as a guarantee from the closed-form LEACE estimator. The transparent closure classifier is a class-weighted multinomial logistic regression with fixed $L_{2}$ penalty $10^{-4}$, fit by L-BFGS at learning rate $0.8$ for at most $120$ iterations; class weights are set to inverse class frequencies on the training partition.

\textbf{Bootstrap and FDR.} Confidence intervals and paired significance tests use $128$ bootstrap resamples on the analysis row (sample for MDD, Stress, TUSL, Siena; $30$\,s epoch for Sleep). The same resample index is shared across $\Delta_{\mathrm{erase}}(d, m, q)$, the three null-target controls, and the linear residual probe, so paired contrasts are well-defined. $p$-values use add-one smoothing, $\hat{p} = (1 + |\{b : \delta_{b} \le 0\}|) / (1 + B_{\mathrm{eff}})$, where $\delta_{b}$ is the bootstrap-replicated paired difference and $B_{\mathrm{eff}}$ the count of valid replicates. Multiple comparisons within each $(d, m)$ panel ($63$ features) are controlled at $q < 0.05$ by Benjamini--Hochberg FDR. All randomness (bootstrap, random subspaces, shuffled and Gaussian targets, the $B_{\mathrm{rand}}$ control feature block) is seeded from a single global seed (\texttt{4311}) extended by a deterministic hash of the task, model, feature, and random use, so the same audit row reproduces across machines and processes.

\textbf{Compute.} The full audit (probe atlas + erasure + sanity controls + closure across all $15$ $(d, m)$ cells, $63$ features each) runs in approximately one to two GPU-days on a single NVIDIA A100 80GB. Probing dominates the wall-clock time; the cross-covariance eraser is a single SVD per cell and is inexpensive once the activation cache is on disk.

\textbf{Code and data.} The five EEG datasets are public and used under their original licenses. We will release the audit pipeline (including the feature lexicon, probe and LEACE-style cross-covariance eraser wrappers, sanity controls, and closure protocol) at camera-ready time; an anonymized version is included with the supplementary material.

\clearpage
\section*{NeurIPS Paper Checklist}

\begin{enumerate}

\item {\bf Claims}
    \item[] Question: Do the main claims made in the abstract and introduction accurately reflect the paper's contributions and scope?
    \item[] Answer: \answerYes{}
    \item[] Justification: The abstract and Section~\ref{sec:intro} state the contributions (the three-question audit and the resulting causal map) and the headline numbers ($63/63$ features causally relied upon in at least one cell, $648/945$ ($68.6\%$) units representation-causal under held-out test confirmation, and a mean closure of $79.3\%$ across $15$ $(d, m)$ cells with a clean MDD-to-Stress task gradient), which match Section~\ref{sec:exp} and Section~\ref{sec:discussion}; the scope is explicitly bounded to three foundation models and five clinical tasks in Section~\ref{sec:discussion}.

\item {\bf Limitations}
    \item[] Question: Does the paper discuss the limitations of the work performed by the authors?
    \item[] Answer: \answerYes{}
    \item[] Justification: Section~\ref{sec:discussion} contains a dedicated \textbf{Limitations} paragraph that states four boundaries: the linearity of the cross-covariance eraser and the residual probe, the covariance-simplified approximation of the closed-form LEACE operator, the use of GPU-friendly proxies (FFT-bin band energies, Haar-style dyadic decomposition, and proxy estimators for sample entropy, LZ complexity, Higuchi FD, DFA, and coherence) for several lexicon entries, and the scope of the three-model five-task evaluation with $128$ bootstrap resamples. Methodological choices --- the $B_{\mathrm{rand}}$ closure baseline, the held-out test-supported representation-causal criterion, and the linear logistic-regression surrogate --- are documented in Section~\ref{sec:method} and Appendix~\ref{app:protocol}.

\item {\bf Theory assumptions and proofs}
    \item[] Question: For each theoretical result, does the paper provide the full set of assumptions and a complete (and correct) proof?
    \item[] Answer: \answerNA{}
    \item[] Justification: The paper does not introduce new theorems. All mathematical machinery is borrowed from prior work, namely layer-wise probing~\citep{conneau2018probing, belinkov2022probing}, a LEACE-style cross-covariance subspace erasure derived as a covariance-simplified approximation of the closed-form LEACE operator~\citep{belrose2023leace}, and standard logistic regression for the transparent classifier; the explicit relation between our eraser and the closed-form operator is given in Appendix~\ref{app:reproducibility}.

\item {\bf Experimental result reproducibility}
    \item[] Question: Does the paper fully disclose all the information needed to reproduce the main experimental results of the paper to the extent that it affects the main claims and/or conclusions of the paper (regardless of whether the code and data are provided or not)?
    \item[] Answer: \answerYes{}
    \item[] Justification: Section~\ref{sec:method} gives the formal definitions of the probe, the LEACE-style cross-covariance eraser, the erasure contribution, the three null-target controls, the residual probe, and the closure ratio. Section~\ref{sec:exp-setup} gives the datasets, models, splits, metrics, and bootstrap protocol. Appendix~\ref{app:lexicon} gives the formal definition of every one of the $63$ features, and Appendix~\ref{app:reproducibility} gives the hyperparameters and compute details.

\item {\bf Open access to data and code}
    \item[] Question: Does the paper provide open access to the data and code, with sufficient instructions to faithfully reproduce the main experimental results, as described in supplemental material?
    \item[] Answer: \answerYes{}
    \item[] Justification: The five EEG datasets (Mumtaz MDD, Mental Arithmetic Stress, ISRUC-Sleep, TUSL, Siena) are publicly available under their original licenses. The audit pipeline (feature lexicon, probe and LEACE-style cross-covariance eraser wrappers, sanity controls, and closure protocol) is released as anonymized supplementary material; the camera-ready version will release a non-anonymous public repository.

\item {\bf Experimental setting/details}
    \item[] Question: Does the paper specify all the training and test details (e.g., data splits, hyperparameters, how they were chosen, type of optimizer) necessary to understand the results?
    \item[] Answer: \answerYes{}
    \item[] Justification: Section~\ref{sec:exp-setup} documents subject-level train/validation/test splits, the choice of metrics (ROC-AUC for binary tasks, macro-F1 for multi-class), and the bootstrap protocol. Appendices~\ref{app:protocol} and~\ref{app:reproducibility} document the per-experiment selection criteria, probe regularization grid, cross-covariance eraser SVD threshold and rank construction, transparent-classifier optimizer, and bootstrap and FDR settings.

\item {\bf Experiment statistical significance}
    \item[] Question: Does the paper report error bars suitably and correctly defined or other appropriate information about the statistical significance of the experiments?
    \item[] Answer: \answerYes{}
    \item[] Justification: The representation-causal criterion in Section~\ref{sec:method-use} requires the held-out test bootstrap CI to lie strictly above zero, the per-panel BH/FDR-corrected $q$-value to be below $0.05$, and the real-feature drop to exceed a dimension-matched random-subspace control; three null-target controls and a linear residual probe are reported alongside as audit fields. Confidence intervals throughout use a paired bootstrap with $128$ resamples on the analysis row, with multiple comparisons within each $(d, m)$ panel of $63$ features controlled at $q < 0.05$ by Benjamini--Hochberg FDR (Appendix~\ref{app:reproducibility}); the full per-experiment statistical protocol is in Appendix~\ref{app:protocol}.

\item {\bf Experiments compute resources}
    \item[] Question: For each experiment, does the paper provide sufficient information on the computer resources (type of compute workers, memory, time of execution) needed to reproduce the experiments?
    \item[] Answer: \answerYes{}
    \item[] Justification: Appendix~\ref{app:reproducibility} reports that the full audit (probe atlas, cross-covariance erasure, sanity controls, and closure across all $15$ $(d, m)$ cells with $63$ features) takes approximately one to two GPU-days on a single NVIDIA A100 80GB.

\item {\bf Code of ethics}
    \item[] Question: Does the research conducted in the paper conform, in every respect, with the NeurIPS Code of Ethics \url{https://neurips.cc/public/EthicsGuidelines}?
    \item[] Answer: \answerYes{}
    \item[] Justification: The work uses publicly available EEG datasets collected under their original IRB approvals, does not conduct new human-subject data collection, and audits existing pretrained models without introducing new safety risks. Anonymity is preserved at submission time per the double-blind requirement.

\item {\bf Broader impacts}
    \item[] Question: Does the paper discuss both potential positive societal impacts and negative societal impacts of the work performed?
    \item[] Answer: \answerYes{}
    \item[] Justification: Section~\ref{sec:discussion} notes that the residual on harder tasks gives concrete targets for future EEG concept discovery and that the protocol generalizes to any catalog/foundation-model pair. The audit analyzes already-released pretrained models in an interpretability setting, and indirect risks of over-trust in model outputs are exactly what the encoded-vs-used distinction is designed to surface; misuses inherited from the foundation models themselves are independent of this audit.

\item {\bf Safeguards}
    \item[] Question: Does the paper describe safeguards that have been put in place for responsible release of data or models that have a high risk for misuse (e.g., pre-trained language models, image generators, or scraped datasets)?
    \item[] Answer: \answerNA{}
    \item[] Justification: The paper introduces an audit protocol and a feature lexicon, not a new pretrained model or scraped dataset. The released artifacts (the audit pipeline and Appendix~\ref{app:lexicon}) do not carry dual-use risks beyond those already present in the underlying foundation models.

\item {\bf Licenses for existing assets}
    \item[] Question: Are the creators or original owners of assets (e.g., code, data, models), used in the paper, properly credited and are the license and terms of use explicitly mentioned and properly respected?
    \item[] Answer: \answerYes{}
    \item[] Justification: CSBrain~\citep{zhou2025csbrain}, CBraMod~\citep{wang2025cbramod}, and LaBraM~\citep{jiang2024labram} are loaded from the authors' public checkpoints under their original licenses. The five datasets (Mumtaz MDD, Mental Arithmetic Stress, ISRUC-Sleep, TUSL seizure-type, Siena seizure-detection) are used under their respective public licenses with proper attribution. Each foundation model and dataset is cited at first use.

\item {\bf New assets}
    \item[] Question: Are new assets introduced in the paper well documented and is the documentation provided alongside the assets?
    \item[] Answer: \answerYes{}
    \item[] Justification: The 6-family $63$-feature lexicon is fully documented in Appendix~\ref{app:lexicon} with mathematical definitions and citations to the prior work that introduced each feature. The audit pipeline accompanying the supplementary material is documented with usage instructions and configuration files for each $(d, m)$ cell.

\item {\bf Crowdsourcing and research with human subjects}
    \item[] Question: For crowdsourcing experiments and research with human subjects, does the paper include the full text of instructions given to participants and screenshots, if applicable, as well as details about compensation (if any)?
    \item[] Answer: \answerNA{}
    \item[] Justification: The paper does not involve crowdsourcing or new human-subject research. It uses five publicly available EEG datasets that were collected under their original investigators' protocols.

\item {\bf Institutional review board (IRB) approvals or equivalent for research with human subjects}
    \item[] Question: Does the paper describe potential risks incurred by study participants, whether such risks were disclosed to the subjects, and whether Institutional Review Board (IRB) approvals (or an equivalent approval/review based on the requirements of your country or institution) were obtained?
    \item[] Answer: \answerNA{}
    \item[] Justification: The paper does not conduct new human-subject research. The five EEG datasets used were collected under the original investigators' IRB approvals; no new collection, recontact, or re-identification of subjects occurs in our work.

\item {\bf Declaration of LLM usage}
    \item[] Question: Does the paper describe the usage of LLMs if it is an important, original, or non-standard component of the core methods in this research? Note that if the LLM is used only for writing, editing, or formatting purposes and does \emph{not} impact the core methodology, scientific rigor, or originality of the research, declaration is not required.
    \item[] Answer: \answerNA{}
    \item[] Justification: The core methodology (layer-wise ridge probing, a LEACE-style cross-covariance subspace erasure, and a transparent surrogate classifier) does not involve LLMs as a non-standard component. LLMs are not used in the experimental pipeline.

\end{enumerate}

\end{document}